\documentclass{article} 
\usepackage{iclr2025_conference,times}

\usepackage{amsmath,amsfonts,bm}

\def\eqref#1{equation~\ref{#1}}

\def\1{\bm{1}}

\DeclareMathAlphabet{\mathsfit}{\encodingdefault}{\sfdefault}{m}{sl}
\SetMathAlphabet{\mathsfit}{bold}{\encodingdefault}{\sfdefault}{bx}{n}

\usepackage{hyperref}
\hypersetup{
    colorlinks=true,       
    citecolor=blue,       
}

\usepackage{url}

\usepackage{xspace}
\usepackage{amsmath}
\usepackage{amsfonts}
\usepackage{subfigure}
\usepackage{graphicx}
\usepackage{booktabs}
\usepackage{url}
\usepackage{tabularx}
\usepackage{bigstrut}
\usepackage{multirow}
\usepackage{color}
\usepackage{enumitem}
\usepackage[normalem]{ulem}
\usepackage{bm}
\usepackage{wrapfig}

\usepackage{tcolorbox}
\tcbuselibrary{raster}
\tcbuselibrary{breakable}
\usepackage{listings}
\usepackage{sourcecodepro}

\makeatletter
\lst@InstallKeywords k{attributes}{attributestyle}\slshape{attributestyle}{}ld
\makeatother

\definecolor{pythonblue}{rgb}{0.16,0.12,0.93}
\definecolor{cppgreen}{rgb}{0.16,0.42,0.16}
\definecolor{promptinsert}{HTML}{bfefff}
\definecolor{compcolor}{HTML}{90EE90}
\definecolor{codehlcolor}{HTML}{ffec8b}
\definecolor{codehlcolor2}{HTML}{ffbbff}
\definecolor{bgcolor}{rgb}{0.95,0.95,0.92}
\definecolor{spblue}{HTML}{00b5ea}

\lstdefinestyle{python}{
    language=Python,
    basicstyle=\fontsize{8}{10}\ttfamily,
    keywordstyle=\color{blue},
    commentstyle=\color{gray},
    stringstyle=\color{black},
    showstringspaces=false,
    breaklines=true,
    breakindent=0pt,
    breakatwhitespace=false,
    escapeinside={(*@}{@*)}
}

\lstdefinestyle{cpp}{
    language=C++,
    basicstyle=\fontsize{8}{10}\ttfamily,
    keywordstyle=\color{blue},
    commentstyle=\color{gray},
    stringstyle=\color{green},
    showstringspaces=false,
    breaklines=true,
    breakindent=0pt,
    breakatwhitespace=false,
    escapeinside={(*@}{@*)}
}

\lstdefinestyle{plain}{
    basicstyle=\fontsize{8}{10}\ttfamily,
    keywordstyle=\color{blue},
    commentstyle=\color{gray},
    stringstyle=\color{green},
    showstringspaces=false,
    breaklines=true,
    breakatwhitespace=false,
    breakindent=0pt,
    escapeinside={(*@}{@*)},
    literate={á}{{\'a}}1 {ã}{{\~a}}1 {é}{{\'e}}1,
}

\lstdefinestyle{demo}{
    basicstyle=\fontsize{7}{8}\ttfamily,
    keywordstyle=\color{blue},
    commentstyle=\color{gray},
    stringstyle=\color{green},
    showstringspaces=false,
    breaklines=true,
    breakatwhitespace=false,
    breakindent=0pt,
    escapeinside={(*@}{@*)},
    literate={á}{{\'a}}1 {ã}{{\~a}}1 {é}{{\'e}}1,
}

\lstdefinestyle{example}{
    basicstyle=\fontsize{8}{10}\ttfamily,
    keywordstyle=\color{spblue}\bfseries\underline,
    commentstyle=\color{gray},
    stringstyle=\color{green},
    showstringspaces=false,
    breaklines=true,
    breakatwhitespace=false,
    breakindent=0pt,
    escapeinside={(*@}{@*)},
    morekeywords={ Question, Answer, Prediction, Results, Explanation },
}

\lstdefinestyle{python2}{
    language=Python,
    basicstyle=\fontsize{8}{10}\ttfamily,
    keywordstyle=\color{blue},
    commentstyle=\color{gray},
    stringstyle=\color{green},
    showstringspaces=false,
    breakatwhitespace=false,
    breaklines=true,
    breakindent=0pt,
    escapeinside={(*@}{@*)}
}

\lstdefinestyle{cpp2}{
    language=C++,
    basicstyle=\fontsize{8}{10}\ttfamily,
    keywordstyle=\color{blue},
    commentstyle=\color{gray},
    stringstyle=\color{green},
    showstringspaces=false,
    breaklines=true,
    breakindent=0pt,
    breakatwhitespace=false,
    escapeinside={(*@}{@*)}
}

\lstdefinestyle{sql}{
    language=SQL,
    basicstyle=\fontsize{8}{10}\ttfamily,
    keywordstyle=\color{blue},
    commentstyle=\color{green},
    stringstyle=\color{black},
    showstringspaces=false,
    breakatwhitespace=false,
    breaklines=true,
    breakindent=0pt,
    escapeinside={(*@}{@*)}
}

\lstdefinestyle{prompt}{
    language=Python,
    basicstyle=\fontsize{8}{10}\ttfamily,
    keywordstyle=\color{blue},
    commentstyle=\color{gray},
    showstringspaces=false,
    breaklines=true,
    keepspaces=true, 
    breakindent=0pt,
    breakatwhitespace=false,
    showspaces=false,   
    escapeinside={(*@}{@*)}
}
\lstdefinestyle{text}{
    basicstyle=\fontsize{8}{10}\ttfamily,
    showstringspaces=false,
    breaklines=true,
    breakatwhitespace=false,
    breakindent=0pt,
    keepspaces=true,
    showspaces=false,   
    escapeinside={(*@}{@*)}
}

\usepackage{algorithm2e}

\newcommand{\method}{\textsc{Speculative RAG}\xspace}
\definecolor{cadmiumgreen}{rgb}{0.0, 0.42, 0.24}

\newcommand{\para}[1]{\textbf{#1}\ }

\title{Speculative RAG: Enhancing Retrieval Augmented Generation through Drafting}

\author{\textbf{Zilong Wang}$^1$\thanks{Work done while the author was a student researcher at Google Cloud AI Research. Correspondence to: Zilong Wang \texttt{<zlwang@ucsd.edu>}, Chen-Yu Lee \texttt{<chenyulee@google.com>}} \quad
        \textbf{Zifeng Wang}$^2$ \quad
        \textbf{Long T. Le}$^2$ \quad
        \textbf{Huaixiu Steven Zheng}$^3$ \\
        \textbf{Swaroop Mishra}$^3$ \quad
        \textbf{Vincent Perot}$^3$ \quad
        \textbf{Yuwei Zhang}$^1$ \quad
        \textbf{Anush Mattapalli}$^4$ \\
        \textbf{Ankur Taly}$^4$ \quad
        \textbf{Jingbo Shang}$^1$ \quad
        \textbf{Chen-Yu Lee}$^2$ \quad
        \textbf{Tomas Pfister}$^2$ \\
  $^1$University of California, San Diego \ \ 
  $^2$Google Cloud AI Research \\ 
  $^3$Google DeepMind \ \
  $^4$Google Cloud AI \ \
}

\iclrfinalcopy 
\begin{document}

\maketitle

\begin{abstract}
  Retrieval augmented generation (RAG) combines the generative abilities of large language models (LLMs) with external knowledge sources to provide more accurate and up-to-date responses. Recent RAG advancements focus on improving retrieval outcomes through iterative LLM refinement or self-critique capabilities acquired through additional instruction tuning of LLMs. In this work, we introduce \method – a framework that leverages a larger generalist LM to efficiently verify multiple RAG drafts produced in parallel by a smaller, distilled specialist LM. Each draft is generated from a distinct subset of retrieved documents, offering diverse perspectives on the evidence \textbf{while reducing input token counts per draft}. This approach enhances comprehension of each subset and mitigates potential \textbf{position bias over long context}. 
Our method accelerates RAG by delegating drafting to the smaller specialist LM, with the larger generalist LM performing a \textbf{single} verification pass over the drafts. Extensive experiments demonstrate that \method achieves \textbf{state-of-the-art performance with reduced latency} on TriviaQA, MuSiQue, PopQA, PubHealth, and ARC-Challenge benchmarks. It notably enhances accuracy by up to 12.97\% while reducing latency by 50.83\% compared to conventional RAG systems on PubHealth.

\end{abstract}

\section{Introduction}

Large language models (LLMs) have demonstrated remarkable success in question answering tasks~\citep{brown2020language,achiam2023gpt,team2023gemini}. Trained on massive datasets, LLMs leverage their extensive parametric memory to generate seemingly plausible responses to user queries~\citep{kojima2022large,kamalloo2023evaluating}. However, when faced with knowledge-intensive questions demanding up-to-date information or obscure facts~\citep{petroni2021kilt}, LLMs struggle with factual inaccuracies and produce hallucinated contents~\citep{huang2023survey}. 

Retrieval Augmented Generation (RAG) has emerged as a promising solution to mitigate these issues. By incorporating information retrieved from an external database into the context~\citep{gao2023retrieval}, RAG effectively reduces factual errors in knowledge-intensive tasks. This approach not only enables easy and efficient access to vast databases but also facilitates timely and accurate knowledge integration.
Due to the inherent limitations in the precision of current dense retrievers and the vastness of knowledge required to answer complex questions~\citep{chen2022deep}, RAG systems typically retrieve multiple documents to ensure the inclusion of all necessary information in the context~\citep{petroni2021kilt}. This practice inevitably increases the length of the input to the LLMs, presenting significant challenges, particularly since encoding lengthy retrieved documents incurs additional latency and require more complex reasoning. 
Recent studies have explored ways to extend the context length limit of LLMs~\citep{ding2023longnet,reid2024gemini,ma2024megalodon}, yet achieving well-grounded reasoning over extended contexts remains an open question~\citep{liu2024lost,li2024long}. Consequently, striking a balance between efficiency and effectiveness in RAG has become a central research question in the literature. 
Existing work on RAG systems primarily concentrates on improving the quality of contextual information in retrieval outcomes, but often neglecting the latency issues associated with these systems~\citep{ma2023query, baek2023knowledge, yan2024corrective, xie2023adaptive, asai2023self, feng2023knowledge}.
These methods typically rely on multiple refinement iterations or customized instruction-tuning for self-critique abilities. Integrating such enhancements into generic LMs requires additional training or increased latency, posing practical challenges in real-world applications.

\begin{figure}[t]
    \centering
    \includegraphics[width=0.9\linewidth]{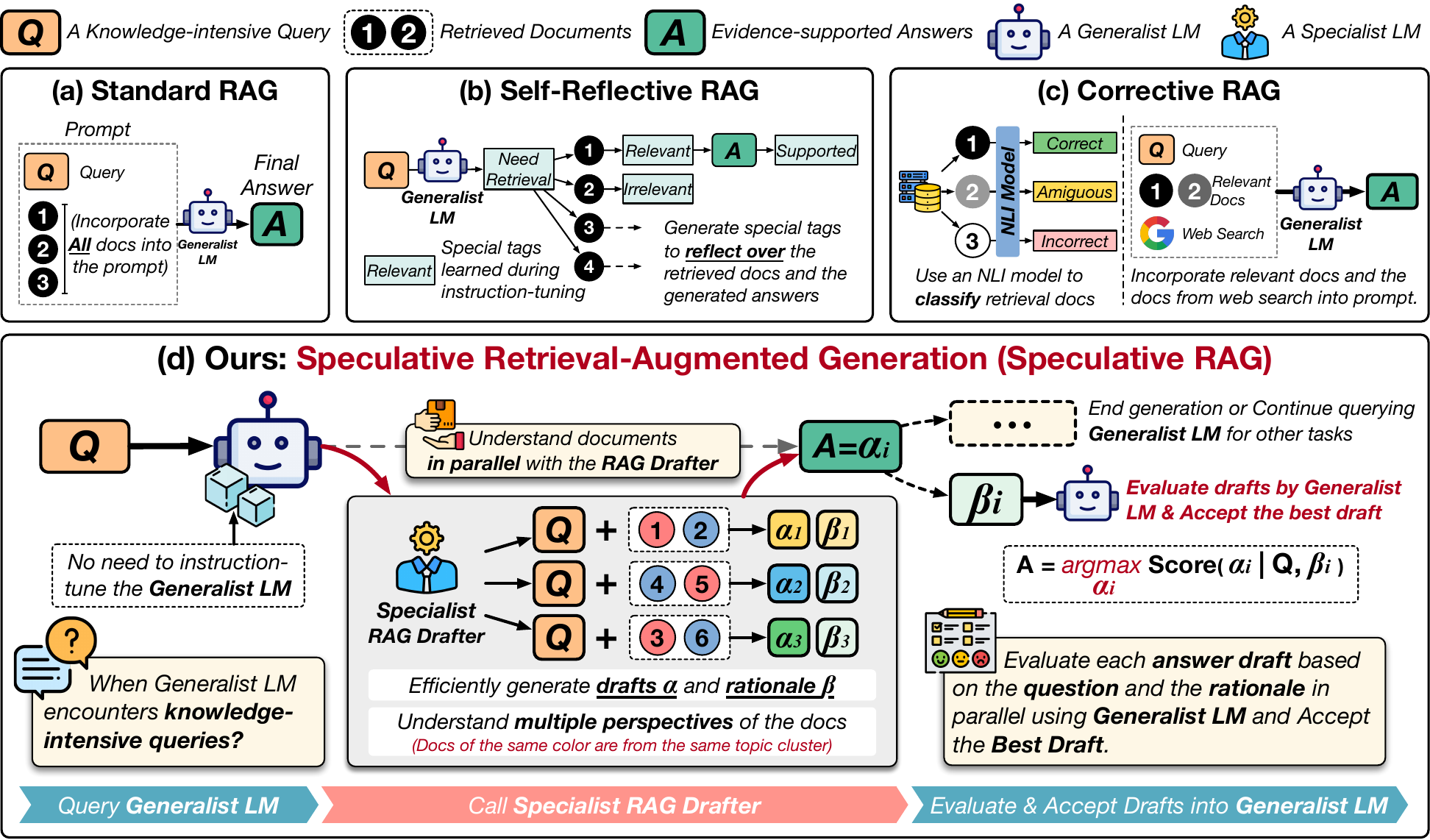}
    \vspace{-3mm}
    \caption{Illustration of different RAG approaches. Given a knowledge-intensive query $Q$ and retrieved documents, (a) Standard RAG incorporates all documents into the prompt, increasing input length and slowing inference; (b) Self-Reflective RAG~\citep{asai2023self} requires specialized instruction-tuning of the general-purpose language model (LM) to generate specific tags for self-reflection; (c) Corrective RAG~\citep{yan2024corrective} employs an external retrieval evaluator to refine document quality, focusing solely on contextual information without enhancing reasoning capabilities; (d) In contrast, our proposed \method leverages a larger generalist LM to efficiently verify multiple RAG drafts produced in parallel by a smaller, specialized LM. Each draft is generated from a distinct subset of retrieved documents, providing diverse perspectives on the evidence while minimizing the number of input tokens per draft.
}
    \vspace{-4mm}
    \label{fig:main_figure}
\end{figure}

To this end, we introduce \method, a RAG framework designed to offload computational burden to a smaller, specialist LM that serves as an efficient and robust RAG module for existing generalist LMs. Inspired by Speculative Decoding~\citep{leviathan2023fast, chen2023accelerating, xia2024unlocking}, which accelerates auto-regressive LM inference by concurrently generating multiple draft tokens with a smaller model and verifying them in parallel with the base model, our approach adapts this concept to RAG.

In \method, we partition retrieved documents into subsets for drafting answer candidates. We cluster the retrieved documents by content similarity and sample one document from each cluster to form a subset, minimizing redundancy and maximizing diversity. These document subsets are then fed to multiple instances of the RAG module, which  generate draft answers with corresponding rationales in parallel. This smaller, specialized RAG module, excels at reasoning over retrieved documents and can rapidly produce accurate responses. Subsequently, the generalist LM bypasses the detailed review of potentially repetitive documents, focusing instead on validating the drafts against the rationales to determine the most accurate answer. We utilize the strong language modeling capabilities of generalist LMs, calculating the conditional generation probability of the answer drafts and rationales as a confidence score. 
Our key contributions are:
\begin{itemize}[leftmargin=*]
\setlength\itemsep{-0.1em}
    \item We introduce a novel RAG framework that employs a smaller specialist RAG drafter to generate high-quality draft answers. Each draft is derived from a distinct subset of retrieved documents, offering diverse perspectives while reducing input token counts per draft. 
    \item The generalist LM, operating with the RAG drafter, requires no additional tuning. It simply verifies and integrates the most promising draft into the final answer. This approach enhances comprehension of each subset and mitigates potential lost-in-the-middle~\citep{liu2024lost} phenomenon. 
    \item Our method significantly accelerates RAG by delegating drafting to the smaller specialist LM, with the larger generalist LM performing a single, unbiased verification pass over the drafts in parallel. Extensive experiments on 5 free-form question-answering and closed-set generation benchmarks demonstrate the superior effectiveness and efficiency of the method.
\end{itemize}

\section{Related Works}

\paragraph{Retrieval Augmented Generation\ }
Retrieval Augmented Generation (RAG) enhances LLMs by retrieving relevant documents from external databases and incorporating them into the generation process~\citep{gao2023retrieval, lewis2020retrieval, khandelwal2020generalization, izacard2021leveraging, luo2023sail,xia2024ground,wang2024rat}. Recent work has primarily focused on enabling LLMs to understand when and what to retrieve~\citep{ma2023query,chen2023densex,jiang2023active,schick2024toolformer}, or designing approaches to better utilize contexts~\citep{yu2023chain,yoran2023making,wang2023learning,sarthi2024raptor,baek2023knowledge,xu2023recomp,kim2024sure}. Among them, SAIL~\citep{luo2023sail} fine-tunes a pre-trained LLM on web search data to filter irrelevant contents. Self-Reflective RAG~\citep{asai2023self} introduces reflection tokens to guide retrieval and annotation in instruction-tuning datasets. However, both approaches require additional instruction-tuning of generic LLMs, which is resource-intensive and may lead to forgetting or over-fitting~\citep{luo2023empirical}. Furthermore, long context with retrieved documents can suffer from computational inefficiency and position bias~\citep{liu2024lost}. Corrective RAG~\citep{yan2024corrective} on the other hand proposes a lightweight retrieval evaluator, but it lacks the capability for high-level reasoning. In contrast, our proposed \method addresses these limitations by leveraging a smaller RAG drafter model to efficiently understand diverse perspectives in retrieval results and generate drafts for the generalist LMs to verify and integrate.

\vspace{-2mm}
\paragraph{Speculative Decoding}
Speculative decoding~\citep{stern2018blockwise,xia2023speculative,chen2023accelerating,leviathan2023fast,xia2024unlocking} aims to reduce auto-regressive decoding latency through a draft-then-verify paradigm. This involves drafting multiple future tokens with a small model and verifying them in parallel with the target model~\citep{xia2024unlocking}. The draft model is typically either an independent model from the same series~\citep{leviathan2023fast,chen2023accelerating} or the target model itself~\citep{zhang2023draft,cai2024medusa}. 
Our approach extends this concept from token-level drafting to answer-level drafting. In contrast to traditional verification criteria~\citep{stern2018blockwise,xia2023speculative,leviathan2023fast,chen2023accelerating,miao2024specinfer}, which accept or reject tokens based on their generation probabilities, we leverage language modeling objectives to directly assess the confidence of entire answer drafts.


\section{Speculative Retrieval Augmented Generation through Drafting}

\para{Problem Formulation}
In knowledge intensive tasks, each entry can be represented as $(Q, D, A)$, where $Q$ is a question or statement that requires additional knowledge; $D = \{d_1, ...,d_n\}$ is a set of $n$ documents retrieved from the database; $A$ is the expected answer. Particularly, in question answering tasks, $Q$ and $A$ are the question and the expected answer in natural language form; in the statement verification tasks, $Q$ is a statement and $A \in \{\texttt{True}, \texttt{False}\}$ is a Boolean value indicating the statement's correctness; in the multiple choice tasks, $Q$ is a question with a few options and $A\in \{\texttt{A},\texttt{B},\texttt{C},...\}$ is the index of the correct answer. The objective of a RAG system is to generate a fluent response containing the expected answer or select the expected answer from the provided options based on the context provided by the retrieved supporting documents. 

\subsection{Overview}\label{sec:overview}
We introduce Speculative Retrieval Augmented Generation (\method), as illustrated in Figure~\ref{fig:main_figure}. We aim at enhancing the reasoning ability of LLMs over retrieved documents without compromising processing speed. 
Instead of relying on brute-force parameter scaling or instruction-tuning an entire LM to handle knowledge-intensive tasks, we propose a divide-and-conquer approach. We utilize \textbf{a smaller specialist LM}, \textbf{the RAG drafter}, to rapidly generate multiple answer drafts based on retrieved results. Then, \textbf{a larger generalist LM}, \textbf{the RAG verifier}, assesses these drafts, selects the best one based on its rationale, and integrates it into the generation results.

\vspace{-3mm}
\SetKwComment{Comment}{$\triangleright$ }{}
\SetKwRepeat{Do}{do}{while}
\SetAlgoLined
\RestyleAlgo{ruled}
\LinesNumbered
\newcommand\mycommentfont[1]{\scriptsize\text{{#1}}}
\SetCommentSty{mycommentfont}
\SetAlCapNameFnt{\small}
\SetAlCapFnt{\small}
\SetKwBlock{Parallel}{
for $\bm{\delta}_j\in \Delta$ do \textit{in parallel}
\Comment*[f]{Process $m$ subsets in parallel.}
}{end}
{
\setlength{\algomargin}{1.em}
\begin{algorithm}
\small
\caption{\method}\label{algo:specrag_overview}
\DontPrintSemicolon
  \KwData{$(Q, D=\{d_i\}_{i}^n)$ is the question and $n$ retrieved documents; $m$  subsets, each containing $k$ documents, are sampled from $D$; $k$ also corresponds to the number of clusters during clustering.
  }
  \KwResult{$\hat{A}$ is the predicted answer to the question.}
  \SetKwFunction{FMain}{Speculative RAG}
  \SetKwProg{Pn}{Function}{:}{\KwRet $\hat{A}$}
  \Pn{\FMain{$Q$, $D$, $m$, $k$}}{
    $\{\bm{c}_1, \bm{c}_2,...,\bm{c}_k\} \xleftarrow{\text{K-Means}} \mathcal{C}({d_1,...,d_n} | Q)$
    \Comment*[f]{Cluster the documents into $k$ groups using an embedding model $\mathcal{C}$.}\label{algoline:cluster}\\
    $\Delta \gets \{\}$\\
    \Repeat{
        $|\Delta| = m$ \Comment*[f]{Repeat the sampling until there are $m$ unique subsets in total.}
    }{
        $\bm{\delta}_j \gets \{\}$\label{algoline:sample_init}
        \Comment*[f]{Construct a subset of the retrieved documents $\bm{\delta}_j$}\\
        \For{$\bm{c}_i \in \{\bm{c}_1,...,\bm{c}_k\}$}{
            $\bm{\delta}_j = \bm{\delta}_j \cup \{\texttt{random.sample}(\bm{c}_i)\}$
            \Comment*[f]{Sample one document from each cluster $\bm{c}_i$ into subset $\bm{\delta}_j$.}
        }\label{algoline:sample_end}
        $\Delta = \Delta \cup \{\bm{\delta}_j\}$
    }
    \Parallel{
        $\alpha_j, \beta_j \gets \mathcal{M}_\text{Drafter}\texttt{.generate}(Q, \bm{\delta}_j)$
        \Comment*[f]{Generate the \textbf{draft} $\alpha$ and \textbf{rationale} $\beta$ with $\mathcal{M}_\text{Drafter}$.}\label{algoline:generate_draft}\\
        
        $\rho_j \gets \mathcal{M}_\text{Verifier}\texttt{.score}(\alpha_j | Q, \beta_j)$
        \Comment*[f]{Compute the confidence score $\rho$ with $\mathcal{M}_\text{Verifier}$.}\label{algoline:confidence_score}\\
    }{}
    $\hat{A} \gets \arg\max_{\alpha_j}\rho_j $
    \Comment*[f]{Select the one with the highest score as the final answer.}\label{algoline:select_best_score}\\
  }
\end{algorithm}
}
\vspace{-3mm}

Specifically, as shown in Algorithm~\ref{algo:specrag_overview}, we first cluster the retrieved documents with regard to their relation to the posed question, where each cluster represents one perspective in the retrieval results (Line~\ref{algoline:cluster}). Then we sample one document from each cluster into a subset so the documents in this subset covers the multiple perspectives in the retrieval results. We aim at minimizing redundancy and increase the diversity of the documents (Line~\ref{algoline:sample_init} to \ref{algoline:sample_end}).
We denote one subset as $\bm{\delta} \subset D$ that contains retrieved documents with diverse contents and multiple perspectives in the retrieval results. Then, we distribute each subset $\bm{\delta}$ to a RAG drafter endpoint $\mathcal{M}_\text{Drafter}$ with the posed question $Q$ to generate the answer draft $\alpha$ and the rationale $\beta$ in parallel (Line~\ref{algoline:generate_draft}). The RAG drafter is instruction-tuned to be a specialist in understanding the retrieved documents and produce rationales that are faithful to the input documents. It is smaller than generalist LMs, and its parallel processing further ensures high efficiency. For each draft-rationale pair $(\alpha, \beta)$ from $\mathcal{M}_\text{Drafter}$, we compute a confidence score with the generalist LM $\mathcal{M}_\text{Verifier}$ based on the question $Q$ and corresponding rationale $\beta$ (Line~\ref{algoline:confidence_score}). It is worth mentioning that $\mathcal{M}_\text{Verifier}$ does not need to be instruction-tuned since we leverage its language modeling ability already learned during pre-training. Meanwhile, $\mathcal{M}_\text{Verifier}$ can verify the drafts based on the informative rationale provided by $\mathcal{M}_\text{Drafter}$ instead of processing tedious or possibly redundant retrieved documents.
Finally, we select the answer draft with the highest confidence score as the final answer and integrate it into the generation results of the generalist LM (Line~\ref{algoline:select_best_score}).

\subsection{Specialist RAG Drafter}\label{sec:drafter}

Instead of tuning a large generalist LM for the RAG scenario, we leverage a smaller specialist LM, $\mathcal{M}_\text{Drafter}$, to understand retrieved documents. $\mathcal{M}_\text{Drafter}$ is specialized in answering the given question based on the supporting documents and not expected to cope with general problems. It serves as a RAG module for the generalist LMs when solving knowledge-intensive tasks. We train $\mathcal{M}_\text{Drafter}$ to generate both the answer draft and the rationale to better understand the contextual documents.

\para{Instruction Tuning} Given a triplet $(Q, A, D)$, where $Q$ is a general query, $A$ is the response, and $D$ is a retrieved supporting document, we augment it with the rationale of the response $A$ based on the document $D$. We denote the rationale as $E$ which extracts essential information from the document and explains why the response is reasonable to the query concisely~\citep{hsieh2023distilling} so it is of shorter length and delivers information coherent with the original document. We leverage relatively strong LMs to automatically synthesize the rationale $E$ for each triplet. Specifically, we directly query the strong LM to understand the knowledge from the document and provide the intermediate rationale between the instruction and response. Refer to Appendix~\ref{appendix:gemini-prompt} for detailed prompts. After generating the rationale, we finetune a pre-trained LM using the standard language modeling objective, maximizing the likelihood: $\mathbb{E}_{(Q,A,D,E)} \log P_{\mathcal{M}_\text{Drafter}}(A, E \mid Q, D)$, 
where $(Q,A,D,E)$ is an augmented entry in the dataset; $P_{\mathcal{M}_\text{Drafter}}(A, E \mid Q, D)$ is the probability of generating the response and rationale based on the query and document.
We use this instruction-tuned model as the specialist RAG drafter which learns to generate a well-grounded response and rationale given the query and relevant documents.

\para{Multi-Perspective Sampling} For each knowledge-intensive question, we retrieve a set of documents from the database using the posed question as the retrieval query. These documents may contain diverse content due to the ambiguity inherent in the query. To minimize redundancy and enhance diversity of the document subsets used for generating answer drafts, we employ a multi-perspective sampling strategy. We first cluster the documents into a few topics using an instruction-aware embedding model~\citep{peng2024answer} and the K-Means clustering~\citep{jin2011k}.
\begin{align*}
    & \small{\texttt{emb}(d_1),...,\texttt{emb}(d_n) = \mathcal{E}(d_1,...,d_n|Q)} \\
      \small{\{\bm{c}_1} & \small{, ..., \bm{c}_k\} = \texttt{K-Means}(\texttt{emb}(d_1),...,\texttt{emb}(d_n))} \\
    & \small{\bm{\delta} = \left\{ \texttt{random.sample}(\bm{c}) \text{ for } \bm{c} \in \{\bm{c}_i\}_1^k\right\}}
\end{align*}
\vspace{-4mm}

where $\mathcal{E}$ is an instruction-aware embedding model which embeds a string with regard to a provided instruction (the posed question $Q$); $\texttt{emb}(d_i)$ is the embedding for the retrieved document $d_i$; $\bm{c}_j$ is a cluster of retrieved documents with similar topics and contents; $k$ is a hyper-parameter that controls the number of clusters. We sample one document from each cluster into a document subset $\bm{\delta}$ so each subset contains $k$ documents of diverse contents. In total, we construct $m$ subsets for parallel inference with the RAG drafter.

\para{RAG Drafting} We run $\mathcal{M}_\text{Drafter}$ over the $m$ document subsets and produce corresponding answer drafts. Refer to Appendix~\ref{appendix:rag-prompt} for detailed prompt. We incorporate each document subset into the prompt and query $\mathcal{M}_\text{Drafter}$ for responses. We obtain $m$ drafts as the answer candidates and each draft is grounded based on the multiple perspectives in the retrieval results. Specifically, given a document subset $\bm{\delta}_j=\{d_{j_1},..,d_{j_k}\}$, we query $\mathcal{M}_\text{Drafter}$ in parallel with the following prompt for the answer draft and rationale: 
$Q, d_{j_1},...,d_{j_k} \to \alpha_j, \beta_j$,
where the prompt contains the posed question $Q$ along with the document subset; the generation result contains the answer draft $\alpha$ and the rationale $\beta$. We denote the conditional generation probability as $\rho_{\text{Draft},j} = P(\beta_j | Q, d_{j_1},...,d_{j_k}) + P( \alpha_j | Q, d_{j_1},...,d_{j_k}, \beta_j)$, which measures the reliability of generating rationales and the confidence in producing answer drafts.


\subsection{Generalist RAG Verifier}\label{sec:verifier}
After generating drafts and the rationale from the RAG drafter $\mathcal{M}_\text{Drafter}$, we evaluate them by a generalist LM $\mathcal{M}_\text{Verifier}$ to filter out the less reliable drafts and select the best answer. The generalist LM can be any off-the-shelf pre-trained LM. We only consider the draft-rationale pair $(\alpha, \beta)$ and skip the tedious and redundant retrieval results. We resort to the language modeling ability of the generalist LM to rank and select the draft-rationale pairs.

\para{Evaluation Scores} 
First, we calculate the \textbf{self-consistency score} by determining the conditional probability of generating a draft-rationale pair given the question, $\rho_\text{Self-contain} = P(\alpha, \beta | Q)$. This score helps assess whether the draft and rationale are self-consistent in the context of the question. Given the characteristics of language modeling, a self-consistent draft-rationale pair is expected to yield a higher probability.
Furthermore, we incorporate a self-reflection statement $R$ that prompts $\mathcal{M}_\text{Verifier}$ to assess the reliability of an answer draft (e.g. ``Do you think the rationale supports the answer, yes or no?''). We define the \textbf{self-reflection score} as $\rho_\text{Self-reflect} = P(\texttt{"Yes"} | Q, \alpha, \beta, R)$ where we compute the conditional probability of the positive answer (\texttt{"Yes"}) to the self-reflection statement.

\para{Computation Method} We can efficiently compute the self-consistency and self-reflection scores within one forward pass of $\mathcal{M}_\text{Verifier}$. Given a question $Q$ and a draft-rationale pair $(\alpha, \beta)$, we construct a prompt $[Q, \alpha, \beta, R, \texttt{"Yes"}]$, where $R$ is the self-reflection statement. We encode the prompt with $\mathcal{M}_\text{Verifier}$, and acquire the probability of each token conditioned on the previous tokens $P(t_i | t_{<i})$. We leverage this auto-regressive feature and aggregate the probability of the relevant tokens to compute the self-consistent score $\rho_\text{Self-contain}$ and self-reflection score $\rho_\text{Self-reflect}$.
\begin{align*}
\small
    \underrightarrow{ Q, \overbrace{\alpha, \beta}^{\rho_\text{SC}}, R, 
    \overbrace{\texttt{"Yes"}}^{\rho_\text{SR}}} \  \Rightarrow
    \begin{cases}
    \rho_\text{SC} = \prod_{t_i\in \alpha} P(t_i|t_{<i}) \cdot \prod_{t_i\in \beta} P(t_i|t_{<i})\\ \rho_\text{SR} = \prod_{t_i\in \texttt{"Yes"}} P(t_i|t_{<i})
    \end{cases}
\end{align*}
Finally, we produce the final score, $\rho_j = \rho_{\text{Draft},j} \cdot \rho_{\text{SC},j} \cdot \rho_{\text{SR},j}$, and then select the most reliable answer as the final answer to the question $\hat{A} = \arg\max_{\alpha_j} \rho_j$.

\section{Experiments}
We evaluate our proposed \method on five public retrieval augmented generation benchmarks: TriviaQA (unfiltered)~\citep{joshi2017triviaqa}, MuSiQue~\citep{trivedi2022musique}, PopQA~\citep{mallen2023not}, PubHealth~\citep{zhang2023interpretable}, and ARC-Challenge~\citep{clark2018think}. We provide representative examples for case study in Appendix~\ref{appendix:case-study}.
TriviaQA, MuSiQue, PopQA are challenging open-domain question answering datasets where RAG systems are required to answer questions on factual knowledge. TriviaQA and PopQA typically require one accurate piece of evidence from the documents, whereas MuSiQue demands multiple documents to construct a multi-hop reasoning chain. More detailed experiments on multi-hop reasoning can be found in Appendix~\ref{sec:multihop}.
Following previous works~\citep{guu2020retrieval, asai2023self, yan2024corrective}, we evaluate performance of the free-form generation based on whether gold answers are contained within the generated response or not. PubHealth and ARC-Challenge are closed-set generation datasets. PubHealth is a dataset of medical claims spanning a variety of biomedical subjects and it requires the RAG system to verify a given claim based on the retrieved documents. ARC-Challenge introduces a multi-choice question answering dataset, composed of science exam questions from grade 3 to grade 9. For closed-set generation tasks, we use accuracy to evaluate whether the generated answers match the ground truth.

\subsection{Baselines}


\paragraph{Standard RAG} 
For standard RAG, we incorporate all the retrieved documents into the prompt as contextual information. Refer to Appendix~\ref{appendix:baseline-prompt} for detailed prompts. We run standard RAG experiments on off-the-shelf LLMs including
Mistral$_\text{7B}$, Mistral-Instruct$_\text{7B}$~\citep{jiang2023mistral}, Mixtral$_\text{8x7B}$, Mixtral-Instruct$_\text{8x7B}$~\citep{jiang2024mixtral}, and Alpaca$_\text{7B}$~\citep{dubois2024alpacafarm}. We also include the performance of Toolformer~\citep{schick2024toolformer} and SAIL~\citep{luo2023sail} which are originally reported from \citet{asai2023self}. Toolformer$_\text{7B}$ is an LM instruction-tuned to use tools including a search engine, and SAIL$_\text{7B}$ is an LM instruction-tuned on the Alpaca instruction tuning set augmented with search results from different sources such as DuckDuckGo and Wikipedia.

% ...

\paragraph{Self-Reflective RAG and Corrective RAG}
Self-Reflective RAG (Self-RAG) ~\citep{asai2023self} and Corrective RAG (CRAG) ~\citep{yan2024corrective} are more advanced RAG systems that enhances the quality of contextual information in the retrieval results. 
CRAG introduces an external evaluator to assess the quality of retrieved documents, and to refine them before the response generation. Self-RAG instruction-tunes an LM to generate special self-refection tags. These tags guides the LM to dynamically retrieve documents when necessary, critique the retrieved documents relevance before generating responses. Self-CRAG is to apply the Self-RAG approach on the refined documents of CRAG. We adopt the same backbone LLMs across all methods as our proposed \method for fair comparisons.


\subsection{Experiment Settings}

In our experiments, we utilize Mistral$_\text{7B}$ (v0.1) as our base LM for the RAG drafter. For RAG verifier, we employ either Mistral$_\text{7B}$ (v0.1) or Mixtral$_\text{8x7B}$ (v0.1) without any fine-tuning, denoted as $\mathcal{M}_\text{Verifier-7B}$ or $\mathcal{M}_\text{Verifier-8x7B}$. We pre-compute embeddings of retrieved documents using a lightweight instruction-aware embedding model InBedder$_\text{Roberta}$~\citep{peng2024answer} as part of the retrieval process. Inference is conducted using the vLLM framework~\citep{kwon2023efficient} with greedy decoding (temperature = 0). We adopt the same experiment settings from \citet{asai2023self} and include a more challenging benchmark, MuSiQue~\citep{trivedi2022musique}. Our focus is on RAG reasoning rather than evidence citation, so we omit the other two long-form generation benchmarks, Biography~\citep{min2023factscore} and ALCE-ASQA~\citep{gao2023enabling}.
On TriviaQA, PopQA, PubHealth, and ARC-Challenge, we retrieve top 10 documents and generate 5 drafts per query ($m=5$), with each draft based on a subset of 2 documents ($k=2$). For MuSiQue, we retrieve top 15 documents and generate 10 drafts for each query ($m=10$), each using a subset of $6$ documents due to more complex reasoning.  
Further details regarding instruction-tuning can be found in Appendix~\ref{appendix:setting}.

\subsection{Main Results}
\begin{table}[t]
  \centering
  \caption{Retrieval augmentation generation results on TriviaQA, MuSiQue, PopQA, PubHealth, and ARC-Challenge (ARC-C). ($^\ast$We use the RAG drafter's generation probability $\rho_\text{Draft}$ as the confidence score for selecting drafts when we use it alone; $^\dagger$ indicates numbers reported in \citet{asai2023self}; $-$ denotes numbers that are not reported by the original papers or are not applicable; $^\ddagger$we use Mistral$_\text{7B}$ or Mixtral$_\text{8x7B}$ as the RAG verifier, and denote them as $\mathcal{M}_\text{Verifier-7B}$ or $\mathcal{M}_\text{Verifier-8x7B}$.)}
\small
  \resizebox{0.9\linewidth}{!}{
    \setlength{\tabcolsep}{1.3mm}{
\begin{tabular}{lccccc}
\toprule
\multirow{2}[4]{*}{\textbf{RAG Method}} & \multicolumn{3}{c}{\textbf{Free-form}} & \multicolumn{2}{c}{\textbf{Closed-set}} \\
\cmidrule(lr){2-4} \cmidrule(lr){5-6}
& \textbf{TriviaQA} & \textbf{MuSiQue} & \textbf{PopQA} & \textbf{PubHealth} & \textbf{ARC-C} \\
\midrule
\multicolumn{5}{l}{\textit{Standard RAG}} \\
\quad Mistral$_\text{7B}$~\citep{jiang2023mistral} & 54.15 & 16.71  & 31.38   & 34.85 & 42.75 \\
\quad Mixtral$_\text{8x7B}$~\citep{jiang2024mixtral} & 59.85 & 19.16 & 34.02     & 37.08 & 48.72 \\

\quad Mistral-Instruct$_\text{7B}$~\citep{jiang2023mistral} & 67.11 &  17.99 & 42.17 & 42.15 & 47.70 \\
\quad Mixtral-Instruct$_\text{8x7B}$~\citep{jiang2024mixtral} & 73.91 & 29.42 & 53.68  & 63.63 & 78.41 \\
\quad Alpaca$_\text{7B}$~\citep{dubois2024alpacafarm}$^\dagger$ & 64.1  & -   & 46.7  & 40.2  & 48.1 \\
\quad Toolformer$_\text{6B}$~\citep{schick2024toolformer}$^\dagger$ & 48.8  & -  & -   & -     & - \\
\quad SAIL$_\text{7B}$~\citep{luo2023sail}$^\dagger$ & -     & -   & -  & 69.2  & 48.4 \\
\midrule
\multicolumn{5}{l}{\textit{Self-Reflective RAG \& Corrective RAG}} \\
\quad CRAG$_\text{Mistral-7B}$~\citep{yan2024corrective} & 59.03 & -  & 49.46   & 59.04 & 74.87 \\
\quad Self-RAG$_\text{Mistral-7B}$~\citep{asai2023self} & 64.84 & 21.72 & 52.68 & 72.44 & 74.91 \\
\quad Self-CRAG$_\text{Mistral-7B}$~\citep{yan2024corrective} & 65.43   & -  & 56.11   & 72.85 & 75.26 \\
\midrule
\multicolumn{5}{l}{\textit{\textbf{Our Speculative RAG}}} \\

\quad $\mathcal{M}_\text{Drafter-7B}$$^\ast$ & 71.11 & 27.89 & 56.40 & 75.58 & 74.49 \\

\quad $\mathcal{M}_\text{Verifier-7B}$$^\ddagger$ + $\mathcal{M}_\text{Drafter-7B}$
& 73.91 & 31.03 & 56.75 & 75.79 & 76.19 \\

\quad $\mathcal{M}_\text{Verifier-8x7B}$$^\ddagger$ + $\mathcal{M}_\text{Drafter-7B}$ 
& \textbf{74.24} &  \textbf{31.57} & \textbf{57.54} &  \textbf{76.60} &  \textbf{80.55} \\

\bottomrule
\end{tabular}%
    }
    }
  \label{tab:main-results}%
\vspace{-4mm}
\end{table}%


We compare \method with standard RAG approaches, as well as the more advanced Self-Reflective RAG and Corrective RAG on five datasets: TriviaQA, MuSiQue, PopQA, PubHealth, and ARC-Challenge. We report the performance of $\mathcal{M}_\text{Drafter-7B}$ when used alone or paired with the RAG verifier (e.g. $\mathcal{M}_\text{Verifier-7B}$, $\mathcal{M}_\text{Verifier-8x7B}$).
Following prior work~\citep{asai2023self,yan2024corrective}, we report accuracy as the performance metric.

\vspace{-2mm}
\paragraph{Superior Performance over Baselines} Table~\ref{tab:main-results} demonstrates that our method consistently outperforms all baselines across all five benchmarks. Particularly, $\mathcal{M}_\text{Verifier-8x7B}$ + $\mathcal{M}_\text{Drafter-7B}$ surpasses the most competitive standard RAG model, Mixtral-Instruct$_\text{8x7B}$, by {0.33\%} on TriviaQA, {2.15\%} on MuSiQue, 3.86\% on PopQA, {12.97\%} on PubHealth, and {2.14\%} on ARC-Challenge. With a comparable number of instruction-tuned parameters, $\mathcal{M}_\text{Verifier-7B}$ + $\mathcal{M}_\text{Drafter-7B}$ outperforms all Self-Reflective and Corrective RAG methods, and $\mathcal{M}_\text{Drafter}$ alone surpasses the baselines in most settings. 

\vspace{-2mm}
\paragraph{Effective Instruction Tuning for RAG Drafter} Our instruction tuning is effective in enhancing the reasoning ability of the drafter model~\citep{hsieh2023distilling}, as we observe a remarkable performance improvement comparing Mistral$_\text{7B}$ and $\mathcal{M}_\text{Drafter-7B}$. Additionally, we further investigate the performance of $\mathcal{M}_\text{Drafter-7B}$ when we directly feed all documents to the RAG drafter and generate one draft, with detailed results provided in Appendix~\ref{appendix:m1kn}.
Moreover, the performance of Mixtral$_\text{8x7B}$ significantly improves when paired with the instruction-tuned RAG drafter $\mathcal{M}_\text{Drafter-7B}$, showing gains of {14.39\%} on TriviaQA, {12.41\%} on MuSiQue, 23.52\% on PopQA, {39.52\%} on PubHealth, and {31.83\%} on ARC-Challenge. Similar improvements are observed with Mistral$_\text{7B}$ as well. For Mistral$_\text{7B}$, we observed improvements of {19.76\%} on TriviaQA, {14.32\%} on MuSiQue, 25.37\% on PopQA, {40.94\%} on PubHealth, and {33.44\%} on ARC-Challenge. We attribute these improvements to the superior reasoning capabilities of the RAG drafter over the retrieved documents in \method. By minimizing the redundancy in the sampled documents, the RAG drafter generates higher quality answer drafts based on diverse perspectives from the retrieval results. 

\para{Reliable Scoring by RAG Verifier} The reliable draft verification by the generalist LM also contributes to the enhanced performance. The performance improves remarkably comparing $\mathcal{M}_\text{Drafter-7B}$ and $\mathcal{M}_\text{Verifier-7B}$ + $\mathcal{M}_\text{Drafter-7B}$. The instruction-tuned RAG drafter is specialized in generating answer drafts based on the retrieved documents while the language modeling capabilities of generic LMs are leveraged to validate each draft in light of its rationale. This method is both effective and easy to implement, showcasing the effectiveness of this verification approach.

\subsection{Effects of Generated Rationale for Verification}
\begin{figure}[h]
    \centering
    \vspace{-3mm}
    \includegraphics[width=0.75\linewidth]{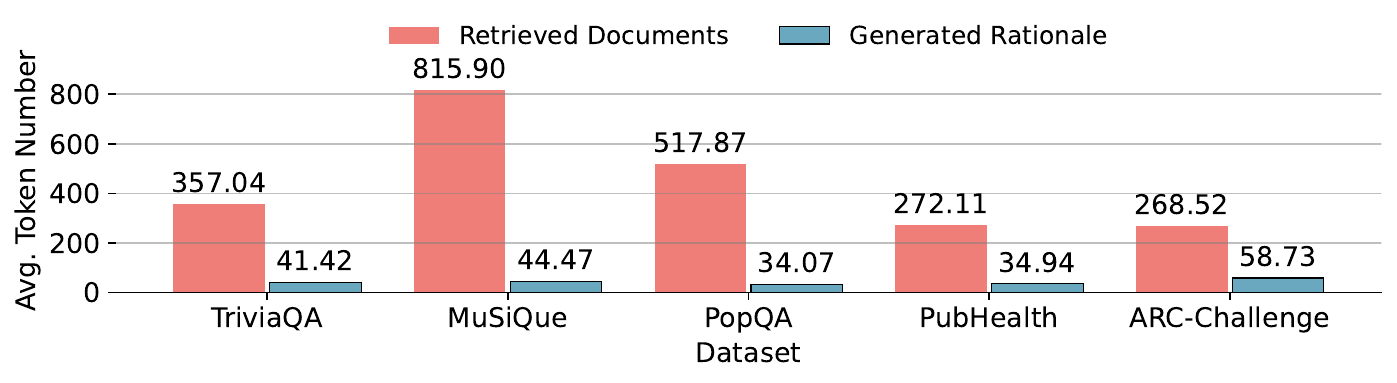}
    \vspace{-4mm}
    \caption{Average number of tokens in the generated rationale and the retrieved documents in TriviaQA, MuSiQue, PopQA, PubHealth, and ARC-Challenge. The generated rationale is of much shorter length than the original retrieved documents.}
    \vspace{-4mm}
    \label{fig:token-num}
\end{figure}

In \method, we utilize the generated rationale $\beta$ from the RAG drafter as an indicator of the trustworthiness of answer drafts $\alpha$. 
\paragraph{Shortened length compared to the retrieved documents.} The rationales highlight relevant points, omit redundant information, and bridge logical gaps between drafts and their supporting documents. We compare the number of tokens in the generated rationale and the retrieved documents, and plot them in Figure~\ref{fig:token-num}. We find that the generated rationale is significantly shorter than the retrieved documents.

\vspace{-3mm}
\begin{table}[h]
  \centering
  \caption{Performance and latency analysis of \method on TriviaQA and PubHealth using $\mathcal{M}_\text{Verifier-8x7B}$ + $\mathcal{M}_\text{Drafter-7B}$. We add the original document subset $\bm{\delta}$ to the context or replace the generated rationale $\beta$ with the original retrieved document subset $\bm{\delta}$ during verification, i.e. we compute the self-containment score as $\rho_\text{Self-contain}=P(\alpha, \bm{\delta}|Q)$ or $\rho_\text{Self-contain}=P(\alpha, \bm{\delta}, \beta|Q)$, and compute the self-reflection score as
  $\rho_\text{Self-reflect}=P(\texttt{"Yes"}|Q, \alpha, \bm{\delta}, R)$ or $\rho_\text{Self-reflect}=P(\texttt{"Yes"}|Q, \alpha, \bm{\delta}, \beta, R)$, where $Q$ is the query; $\alpha$ is the answer draft; $R$ is the self-reflection statement.}
  \small
    \resizebox{0.9\linewidth}{!}{
    \setlength{\tabcolsep}{4.5mm}{
    \begin{tabular}{lllll}
    \toprule
    \multirow{2}[4]{*}{} & \multicolumn{2}{c}{\textbf{TriviaQA}} & \multicolumn{2}{c}{\textbf{PubHealth}} \\
\cmidrule(lr){2-3} \cmidrule(lr){4-5}         & \multicolumn{1}{l}{\textbf{Accuracy (\%)}} & \multicolumn{1}{l}{\textbf{Latency (s)}} & \multicolumn{1}{l}{\textbf{Accuracy (\%)}} & \multicolumn{1}{l}{\textbf{Latency (s)}} \\
    \midrule
    $\mathcal{M}_\text{Verifier-8x7B}$ + $\mathcal{M}_\text{Drafter-7B}$ &       &       &       &  \\
    \quad $\rho = \texttt{Score}(\alpha | Q, \beta)$ &  74.24   &  \textbf{1.93}  &  \textbf{76.60}   &  \textbf{1.17}  \\
    \quad $\rho = \texttt{Score}(\alpha | Q, \bm{\delta})$ & 74.08 \scriptsize{\textcolor{red}{(-0.16)}}  &   2.13 \scriptsize{\textcolor{red}{(+10.36\%)}}    &  76.09 \scriptsize{\textcolor{red}{(-0.51)}}  & 1.31 \scriptsize{\textcolor{red}{(+11.97\%)}}  \\
    \quad $\rho = \texttt{Score}(\alpha | Q, \beta, \bm{\delta})$ &  \textbf{74.32} \scriptsize{\textcolor{cadmiumgreen}{(+0.08)}}  &   2.17 \scriptsize{\textcolor{red}{(+12.44\%)}} &  76.29 \scriptsize{\textcolor{red}{(-0.31)}}  &  1.33 \scriptsize{\textcolor{red}{(+13.68\%)}} \\
    \bottomrule
    \end{tabular}%
    }}
  \label{tab:rationale-ablation}%
  \vspace{-3mm}
\end{table}%

\paragraph{Comparable performance with retrieved documents and lower latency.} To evaluate the effectiveness of the rationales, we create two alternative scoring methods: (a) replacing rationale with retrieved documents ($\rho = \texttt{Score}(\alpha | Q, \bm{\delta})$), or (b) adding retrieved documents to rationale ($\rho = \texttt{Score}(\alpha | Q, \beta, \bm{\delta})$). We compare these alternatives to the scoring method used in \method ($\rho = \texttt{Score}(\alpha | Q, \beta)$) in Table~\ref{tab:rationale-ablation}. 
The results show that incorporating longer retrieved documents does not consistently improve performance and tends to increase latency. This suggest that the generated rationale is already of high quality and serves as an effective bridge between the supporting documents and the generated answer drafts. By leveraging this rationale, we can efficiently verify drafts using a generic LM, leading to accurate final results. We further validate the rationale generation in the instruction-tuning stage. See Appendix~\ref{appendix:rationale-generation} for more details.

\subsection{Latency Analysis with Baselines}
\begin{figure}[t]
    \centering
    \includegraphics[width=0.98\linewidth]{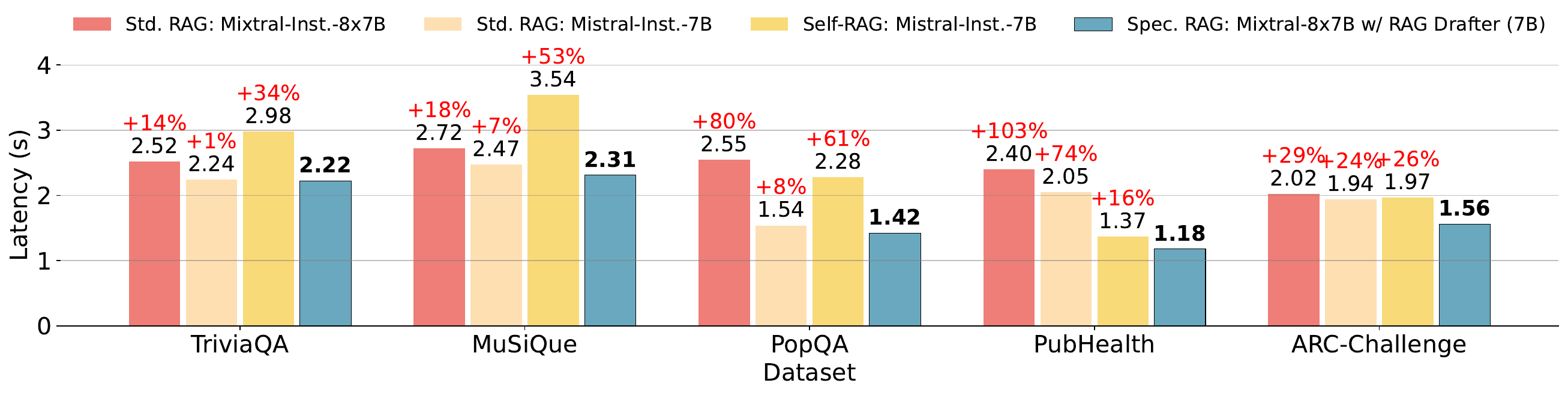}
    \vspace{-3mm}
    \caption{Latency analysis of Standard RAG, Self-RAG, and \method on TriviaQA, MuSiQue, PopQA, PubHealth, and ARC-Challenge. The latency difference between Standard RAG/Self-RAG and \method is highlighted in \textcolor{red}{\textit{red} (+$x$\%)}.
    The latency varies across different datasets due to different retrieved document lengths. \method encodes the retrieved documents in parallel and generates answer drafts with a smaller RAG drafter. This significantly improves the efficiency. }
    \vspace{-2mm}
    \label{fig:latency-bar}
\end{figure}

We analyze the latency of Standard RAG, Self-RAG, and our \method on TriviaQA, MuSiQue, PopQA, PubHealth, and ARC-Challenge. We randomly sample 100 cases from each dataset and report the average time cost for each case, as shown in Figure~\ref{fig:latency-bar}. To simulate real-world application scenarios, we process cases individually without batching.
As representative example, we run $\mathcal{M}_\text{Verifier-8x7B}$ + $\mathcal{M}_\text{Drafter-7B}$ for \method and Mixtral-Instruct$_\text{8x7B}$ for Standard RAG, as these demonstrate the highest performance among competitive baselines (see Table~\ref{tab:main-results}). We also include the analysis for Standard RAG: Mistral-Instruct$_\text{7B}$ and Self-RAG: Mistral-Instruct$_\text{7B}$ in this study.
For \method, we launch 5 endpoints of $\mathcal{M}_\text{Drafter-7B}$ for parallel drafting on TriviaQA, PopQA, PubHealth, and ARC-Challenge. We launch 10 endpoints for MuSiQue due to more drafts. We use tensor parallelism of 4 to fit Mixtral-Instruct$_\text{8x7B}$ into the GPU memory. We use the same tensor parallelism setting for the other methods for a fair comparison.

\paragraph{Reducing processing time while maintaining high performance} As the results demonstrate, \method consistently achieves the lowest latency compared to all other methods. This advantage comes from its utilization of fewer documents needed per draft and parallel drafting.
Particularly, compared with the most competitive baseline, Standard RAG: $\mathcal{M}_\text{Verifier-8x7B}$ + $\mathcal{M}_\text{Drafter-7B}$, our proposed \method reduces latency by up to {11.90\%} on TriviaQA, {15.07\%} on MuSiQue, 44.31\% on PopQA, {50.83\%} on PubHealth, and {22.77\%} on ARC-Challenge. 
Furthermore, a direct comparison between Standard RAG: Mistral-Instruct$_\text{7B}$ and our method reveals that the higher latency of Standard RAG: Mistral-Instruct$_\text{7B}$ is due to its longer context length which contains all retrieved documents. Self-RAG: Mistral-Instruct$_\text{7B}$ also exhibits higher latency due to the generation of longer answers with self-reflection tags and the additional overhead associated with evidence selection.
These findings highlight the advantage of our approach in reducing processing time while maintaining high performance.


\subsection{Ablation Studies}
We conduct ablation studies on the multi-perspective sampling (Section~\ref{sec:drafter}) and the evaluation scores (Section~\ref{sec:verifier}) of \method during the drafting or the verification stages on TriviaQA and PubHealth in Table~\ref{tab:ablation}. We use $\mathcal{M}_\text{Verifier-8x7B}$ + $\mathcal{M}_\text{Drafter-7B}$ as a running configuration. Same as the main results, we report the accuracy as performance metrics.

\para{Diversity and reduced redundancy in retrieval improves draft quality significantly.} In the first set of experiments, we evaluate the impact of  multi-perspective sampling during the drafting. Recall that \method clusters retrieved documents into distinct perspectives and sample one document from each cluster to reduce redundancy for the draft generation. We compare this against two alternative sampling strategies: (1) Random sampling without multi-perspective clustering, where we randomly select a document subset as context, and (2) Always sampling from the same cluster, where we select all documents from a single cluster. Our results indicate that our proposed sampling method yields the best performance thanks to its ability to leverage diverse context. Particularly, it improves the accuracy up to 1.88\% on TriviaQA and 2.23\% on PubHealth. While random sampling without clustering introduces diversity, it is prone to including redundant documents, degrading draft quality. Sampling from the same cluster significantly underperforms due to a lack of diverse perspectives.

\para{Scoring method on self-consistency and self-reflection refines draft quality effectively.}
In the second set of experiments, we examine the scoring method during verification. We remove each of the specific confidence scores, $\rho_\text{Draft}$, $\rho_\text{Self-contain}$, or $\rho_\text{Self-reflect}$ in turn. Performance drops are observed when any score is removed. Particularly, removing $\rho_\text{Draft}$ leads to a minimal decline, 0.19\% on TriviaQA and 1.12\% on PubHealth, likely due to the limited verification capability of the smaller RAG drafter. Removing either $\rho_\text{Self-contain}$ or $\rho_\text{Self-reflect}$ results in similar performance decreases, around 2.0\% on TriviaQA and around 0.8\% on PubHealth, indicating that both self-containment and self-reflection capture different key aspects of reasoning and are crucial during verification. Random selection without verification leads to substantial underperformance, resulting in a performance decline of 5.69\% on TriviaQA and 5.37\% on PubHealth.

\begin{table}[ht]
  \centering
  \caption{Ablation study of \method in the drafting and verification stages on TriviaQA and PubHealth.}
  \small
    \resizebox{0.9\linewidth}{!}{
    \setlength{\tabcolsep}{4.5mm}{
    \begin{tabular}{lll}
    \toprule
          & \textbf{TriviaQA} & \textbf{PubHealth} \\
    \midrule
    $\mathcal{M}_\text{Verifier-8x7B}$ + $\mathcal{M}_\text{Drafter-7B}$  & \textbf{74.24} & \textbf{76.60} \\
    \midrule
    \textit{Drafting Stage} &       &  \\
    \quad Random sampling \textit{w/o} multi-perspective clustering &  
    73.02 \scriptsize{\textcolor{red}{(-1.22)}}  & 
    75.38 \scriptsize{\textcolor{red}{(-1.22)}} \\
    \quad Always sampling from the same perspective cluster & 
    72.36 \scriptsize{\textcolor{red}{(-1.88)}}  & 
    74.37 \scriptsize{\textcolor{red}{(-2.23)}} \\

    \midrule
    \textit{Verification Stage} &       &  \\
    \quad \textit{w/o} $\rho_\text{Draft}$ ($\rho = \rho_\text{Self-contain} \cdot \rho_\text{Self-reflect}$) & 
    74.05 \scriptsize{\textcolor{red}{(-0.19)}} & 
    75.48 \scriptsize{\textcolor{red}{(-1.12)}} \\
    \quad \textit{w/o} $\rho_\text{Self-contain}$ ($\rho = \rho_\text{Draft} \cdot \rho_\text{Self-reflect}$) & 
    72.04 \scriptsize{\textcolor{red}{(-2.20)}} & 
    75.89 \scriptsize{\textcolor{red}{(-0.71)}} \\
    \quad \textit{w/o} $\rho_\text{Self-reflect}$ ($\rho = \rho_\text{Draft} \cdot \rho_\text{Self-contain}$) & 
    72.36 \scriptsize{\textcolor{red}{(-1.88)}} & 
    75.68 \scriptsize{\textcolor{red}{(-0.92)}} \\
    
    \quad Random selection \textit{w/o} verification & 
    68.55 \scriptsize{\textcolor{red}{(-5.69)}} & 
    71.23 \scriptsize{\textcolor{red}{(-5.37)}} \\
    
    \bottomrule
    \end{tabular}%
    }
    }
  \label{tab:ablation}%
\end{table}%

\begin{figure}[ht]
    \centering
    \vspace{-3mm}
    \subfigure[We include 1, 5, 10, 15, 20 drafts and sample 2 supporting documents for each draft.]{%
        \includegraphics[width=0.485\textwidth]{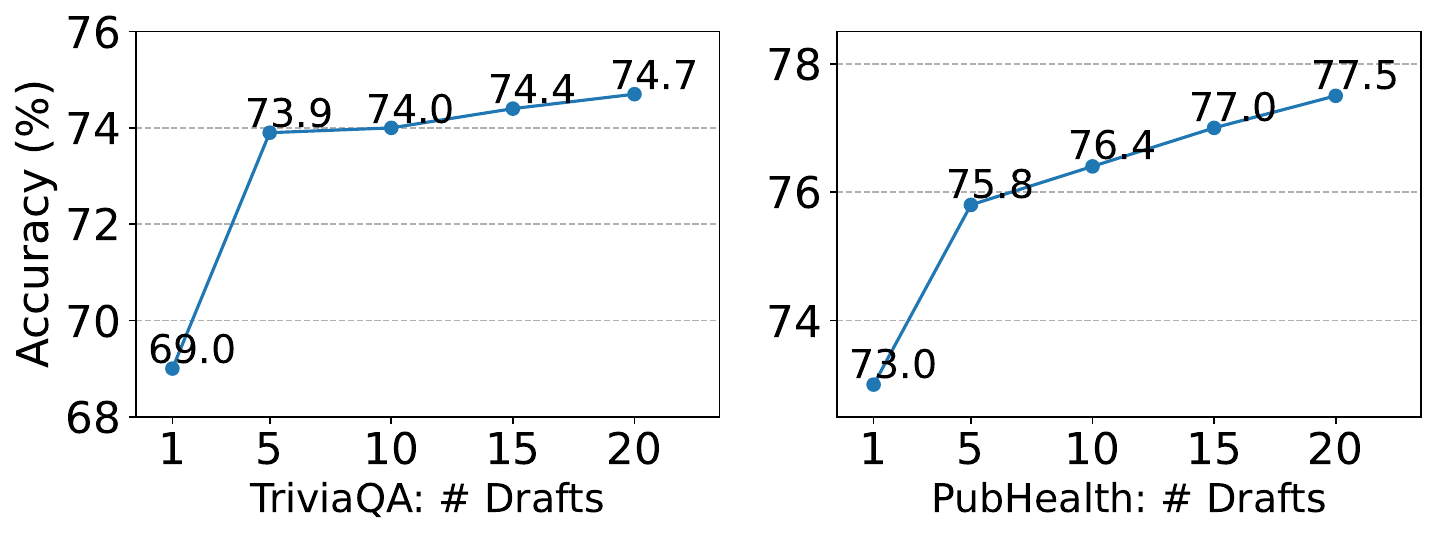}
    }
    \hfill
    \subfigure[We sample 1, 2, 4, 6, 10 supporting documents for each draft and we generate 10 answer drafts.]{%
        \includegraphics[width=0.48\textwidth]{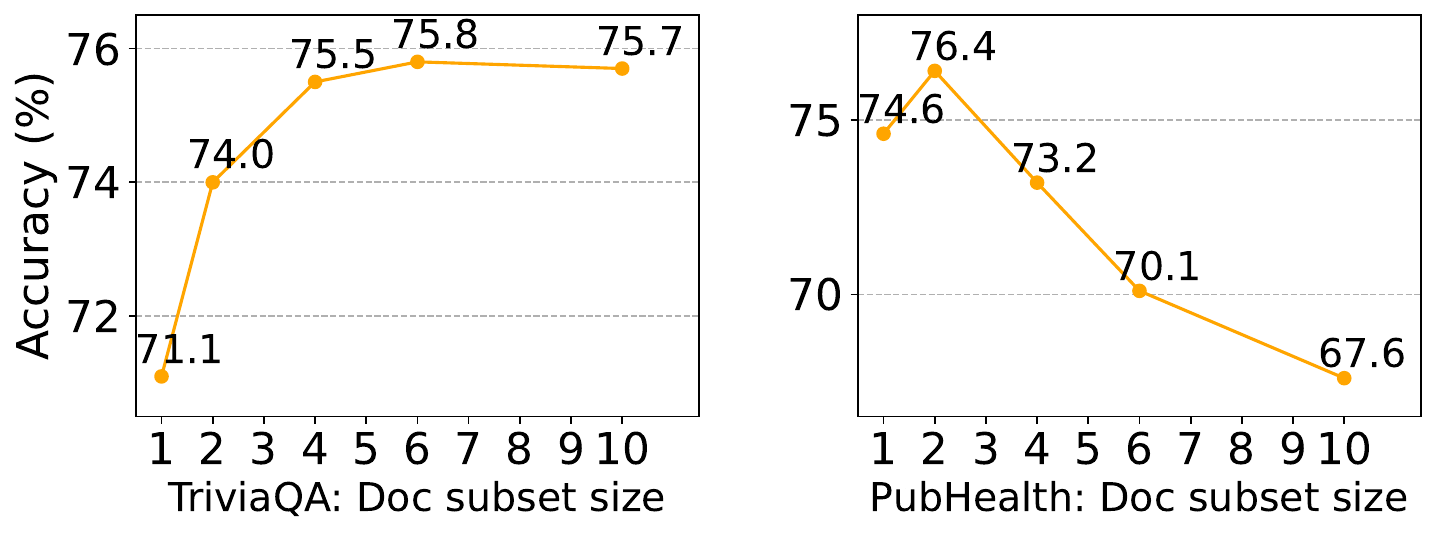}
    }
    \vspace{-3mm}
    \caption{Performance analysis of \method with (a) different numbers of drafts, and (b) different supporting document subset size on TriviaQA and PubHealth.}
    \label{fig:tendency-ablation}
    \vspace{-3mm}
\end{figure}

\subsection{Effects of Draft Number and Document Subset Size}

\paragraph{Increasing the number of drafts improves performance without adding latency.} We investigate the performance of \method under varying numbers of drafts. Using $\mathcal{M}_\text{Verifier-7B}$ + $\mathcal{M}_\text{Drafter-7B}$ with 1, 5, 10, 15, 20 drafts on TriviaQA and PubHealth. We sample two documents as context per draft. The results are illustrated in Figure~\ref{fig:tendency-ablation}(a). Since we retrieve top 10 documents in total, we sample up to 20 drafts in these experiments. The results indicate that incorporating more drafts can further improve performance, likely thanks to higher coverage of diverse perspective of documents. Importantly, in \method, we can launch multiple RAG drafter instances to generate drafts in parallel without additional latency.

\vspace{-3mm}
\paragraph{Increasing the document subset size doesn't always lead to better performance.} We also examine the effect of document subset size. By varying the number of documents (1, 2, 4, 6, or 10) sampled for draft generation on TriviaQA and PubHealth (Figure~\ref{fig:tendency-ablation}(b)), we find that including more documents in the context does not always lead to consistent performance improvement. While TriviaQA queries may benefit from more supporting documents due to their complexity,  $\mathcal{M}_\text{Verifier-7B}$ + $\mathcal{M}_\text{Drafter-7B}$ can surpass Mistral-Instruct$_\text{7B}$ even with a single supporting document per draft. Furthermore, with two or more documents per draft, $\mathcal{M}_\text{Verifier-7B}$ + $\mathcal{M}_\text{Drafter-7B}$ can even surpass Mixtral-Instruct$_\text{8x7B}$. This further demonstrates the effectiveness of our drafting design.


\section{Conclusion}

Our proposed \method decomposes RAG tasks into two separate steps of drafting followed by verification.
\method delegates the heavy lifting of drafting to a small specialized RAG drafter, while verification is done using a large generalist LM. The parallel generation of multiple drafts from diverse document subsets provides high quality answer candidates while reducing input token counts and the potential risk of position-bias-over-long-context, resulting in substantial improvements in both the quality and speed of the final output generation. We demonstrate the effectiveness of \method with accuracy gains up to 12.97\% while reducing latency by 50.83\% compared to conventional RAG systems. \method sheds new light on the potential of collaborative architectures for enhancing RAG performance through task decomposition.





\bibliography{reference}

\begin{thebibliography}{69}
\providecommand{\natexlab}[1]{#1}
\providecommand{\url}[1]{\texttt{#1}}
\expandafter\ifx\csname urlstyle\endcsname\relax
  \providecommand{\doi}[1]{doi: #1}\else
  \providecommand{\doi}{doi: \begingroup \urlstyle{rm}\Url}\fi

\bibitem[Achiam et~al.(2023)Achiam, Adler, Agarwal, Ahmad, Akkaya, Aleman,
  Almeida, Altenschmidt, Altman, Anadkat, et~al.]{achiam2023gpt}
Josh Achiam, Steven Adler, Sandhini Agarwal, Lama Ahmad, Ilge Akkaya,
  Florencia~Leoni Aleman, Diogo Almeida, Janko Altenschmidt, Sam Altman,
  Shyamal Anadkat, et~al.
\newblock Gpt-4 technical report.
\newblock \emph{arXiv preprint arXiv:2303.08774}, 2023.

\bibitem[Asai et~al.(2023)Asai, Wu, Wang, Sil, and Hajishirzi]{asai2023self}
Akari Asai, Zeqiu Wu, Yizhong Wang, Avirup Sil, and Hannaneh Hajishirzi.
\newblock Self-rag: Learning to retrieve, generate, and critique through
  self-reflection.
\newblock \emph{arXiv preprint arXiv:2310.11511}, 2023.

\bibitem[Baek et~al.(2023)Baek, Jeong, Kang, Park, and
  Hwang]{baek2023knowledge}
Jinheon Baek, Soyeong Jeong, Minki Kang, Jong~C Park, and Sung Hwang.
\newblock Knowledge-augmented language model verification.
\newblock In \emph{Proceedings of the 2023 Conference on Empirical Methods in
  Natural Language Processing}, pp.\  1720--1736, 2023.

\bibitem[Brown et~al.(2020)Brown, Mann, Ryder, Subbiah, Kaplan, Dhariwal,
  Neelakantan, Shyam, Sastry, Askell, et~al.]{brown2020language}
Tom Brown, Benjamin Mann, Nick Ryder, Melanie Subbiah, Jared~D Kaplan, Prafulla
  Dhariwal, Arvind Neelakantan, Pranav Shyam, Girish Sastry, Amanda Askell,
  et~al.
\newblock Language models are few-shot learners.
\newblock \emph{Advances in neural information processing systems},
  33:\penalty0 1877--1901, 2020.

\bibitem[Cai et~al.(2024)Cai, Li, Geng, Peng, Lee, Chen, and
  Dao]{cai2024medusa}
Tianle Cai, Yuhong Li, Zhengyang Geng, Hongwu Peng, Jason~D Lee, Deming Chen,
  and Tri Dao.
\newblock Medusa: Simple llm inference acceleration framework with multiple
  decoding heads.
\newblock \emph{arXiv preprint arXiv:2401.10774}, 2024.

\bibitem[Chen et~al.(2023{\natexlab{a}})Chen, Borgeaud, Irving, Lespiau, Sifre,
  and Jumper]{chen2023accelerating}
Charlie Chen, Sebastian Borgeaud, Geoffrey Irving, Jean-Baptiste Lespiau,
  Laurent Sifre, and John Jumper.
\newblock Accelerating large language model decoding with speculative sampling.
\newblock \emph{arXiv preprint arXiv:2302.01318}, 2023{\natexlab{a}}.

\bibitem[Chen et~al.(2023{\natexlab{b}})Chen, Wang, Chen, Yu, Ma, Zhao, Zhang,
  and Yu]{chen2023densex}
Tong Chen, Hongwei Wang, Sihao Chen, Wenhao Yu, Kaixin Ma, Xinran Zhao,
  Hongming Zhang, and Dong Yu.
\newblock Dense x retrieval: What retrieval granularity should we use?
\newblock \emph{arXiv preprint arXiv:2312.06648}, 2023{\natexlab{b}}.

\bibitem[Chen et~al.(2022)Chen, Liu, Wang, Bakker, Georgiou, Fieguth, Liu, and
  Lew]{chen2022deep}
Wei Chen, Yu~Liu, Weiping Wang, Erwin~M Bakker, Theodoros Georgiou, Paul
  Fieguth, Li~Liu, and Michael~S Lew.
\newblock Deep learning for instance retrieval: A survey.
\newblock \emph{IEEE Transactions on Pattern Analysis and Machine
  Intelligence}, 2022.

\bibitem[Clark et~al.(2018)Clark, Cowhey, Etzioni, Khot, Sabharwal, Schoenick,
  and Tafjord]{clark2018think}
Peter Clark, Isaac Cowhey, Oren Etzioni, Tushar Khot, Ashish Sabharwal, Carissa
  Schoenick, and Oyvind Tafjord.
\newblock Think you have solved question answering? try arc, the ai2 reasoning
  challenge.
\newblock \emph{arXiv preprint arXiv:1803.05457}, 2018.

\bibitem[Ding et~al.(2023)Ding, Ma, Dong, Zhang, Huang, Wang, Zheng, and
  Wei]{ding2023longnet}
Jiayu Ding, Shuming Ma, Li~Dong, Xingxing Zhang, Shaohan Huang, Wenhui Wang,
  Nanning Zheng, and Furu Wei.
\newblock Longnet: Scaling transformers to 1,000,000,000 tokens.
\newblock \emph{arXiv preprint arXiv:2307.02486}, 2023.

\bibitem[Dubois et~al.(2024)Dubois, Li, Taori, Zhang, Gulrajani, Ba, Guestrin,
  Liang, and Hashimoto]{dubois2024alpacafarm}
Yann Dubois, Chen~Xuechen Li, Rohan Taori, Tianyi Zhang, Ishaan Gulrajani,
  Jimmy Ba, Carlos Guestrin, Percy~S Liang, and Tatsunori~B Hashimoto.
\newblock Alpacafarm: A simulation framework for methods that learn from human
  feedback.
\newblock \emph{Advances in Neural Information Processing Systems}, 36, 2024.

\bibitem[Feng et~al.(2023)Feng, Shi, Bai, Balachandran, He, and
  Tsvetkov]{feng2023knowledge}
Shangbin Feng, Weijia Shi, Yuyang Bai, Vidhisha Balachandran, Tianxing He, and
  Yulia Tsvetkov.
\newblock Knowledge card: Filling llms' knowledge gaps with plug-in specialized
  language models.
\newblock In \emph{The Twelfth International Conference on Learning
  Representations}, 2023.

\bibitem[Gao et~al.(2023{\natexlab{a}})Gao, Yen, Yu, and Chen]{gao2023enabling}
Tianyu Gao, Howard Yen, Jiatong Yu, and Danqi Chen.
\newblock Enabling large language models to generate text with citations.
\newblock In \emph{Proceedings of the 2023 Conference on Empirical Methods in
  Natural Language Processing}, pp.\  6465--6488, 2023{\natexlab{a}}.

\bibitem[Gao et~al.(2023{\natexlab{b}})Gao, Xiong, Gao, Jia, Pan, Bi, Dai, Sun,
  and Wang]{gao2023retrieval}
Yunfan Gao, Yun Xiong, Xinyu Gao, Kangxiang Jia, Jinliu Pan, Yuxi Bi, Yi~Dai,
  Jiawei Sun, and Haofen Wang.
\newblock Retrieval-augmented generation for large language models: A survey.
\newblock \emph{arXiv preprint arXiv:2312.10997}, 2023{\natexlab{b}}.

\bibitem[Guu et~al.(2020)Guu, Lee, Tung, Pasupat, and Chang]{guu2020retrieval}
Kelvin Guu, Kenton Lee, Zora Tung, Panupong Pasupat, and Mingwei Chang.
\newblock Retrieval augmented language model pre-training.
\newblock In \emph{International conference on machine learning}, pp.\
  3929--3938. PMLR, 2020.

\bibitem[Hsieh et~al.(2023)Hsieh, Li, Yeh, Nakhost, Fujii, Ratner, Krishna,
  Lee, and Pfister]{hsieh2023distilling}
Cheng-Yu Hsieh, Chun-Liang Li, Chih-Kuan Yeh, Hootan Nakhost, Yasuhisa Fujii,
  Alexander Ratner, Ranjay Krishna, Chen-Yu Lee, and Tomas Pfister.
\newblock Distilling step-by-step! outperforming larger language models with
  less training data and smaller model sizes.
\newblock In \emph{ACL}, 2023.

\bibitem[Huang et~al.(2023)Huang, Yu, Ma, Zhong, Feng, Wang, Chen, Peng, Feng,
  Qin, et~al.]{huang2023survey}
Lei Huang, Weijiang Yu, Weitao Ma, Weihong Zhong, Zhangyin Feng, Haotian Wang,
  Qianglong Chen, Weihua Peng, Xiaocheng Feng, Bing Qin, et~al.
\newblock A survey on hallucination in large language models: Principles,
  taxonomy, challenges, and open questions.
\newblock \emph{arXiv preprint arXiv:2311.05232}, 2023.

\bibitem[Izacard \& Grave(2021)Izacard and Grave]{izacard2021leveraging}
Gautier Izacard and Edouard Grave.
\newblock Leveraging passage retrieval with generative models for open domain
  question answering.
\newblock In Paola Merlo, Jorg Tiedemann, and Reut Tsarfaty (eds.),
  \emph{Proceedings of the 16th Conference of the European Chapter of the
  Association for Computational Linguistics: Main Volume}, pp.\  874--880,
  Online, April 2021. Association for Computational Linguistics.
\newblock \doi{10.18653/v1/2021.eacl-main.74}.
\newblock URL \url{https://aclanthology.org/2021.eacl-main.74}.

\bibitem[Izacard et~al.(2021)Izacard, Caron, Hosseini, Riedel, Bojanowski,
  Joulin, and Grave]{izacard2021unsupervised}
Gautier Izacard, Mathilde Caron, Lucas Hosseini, Sebastian Riedel, Piotr
  Bojanowski, Armand Joulin, and Edouard Grave.
\newblock Unsupervised dense information retrieval with contrastive learning.
\newblock \emph{arXiv preprint arXiv:2112.09118}, 2021.

\bibitem[Jiang et~al.(2023{\natexlab{a}})Jiang, Sablayrolles, Mensch, Bamford,
  Chaplot, Casas, Bressand, Lengyel, Lample, Saulnier,
  et~al.]{jiang2023mistral}
Albert~Q Jiang, Alexandre Sablayrolles, Arthur Mensch, Chris Bamford,
  Devendra~Singh Chaplot, Diego de~las Casas, Florian Bressand, Gianna Lengyel,
  Guillaume Lample, Lucile Saulnier, et~al.
\newblock Mistral 7b.
\newblock \emph{arXiv preprint arXiv:2310.06825}, 2023{\natexlab{a}}.

\bibitem[Jiang et~al.(2024)Jiang, Sablayrolles, Roux, Mensch, Savary, Bamford,
  Chaplot, Casas, Hanna, Bressand, et~al.]{jiang2024mixtral}
Albert~Q Jiang, Alexandre Sablayrolles, Antoine Roux, Arthur Mensch, Blanche
  Savary, Chris Bamford, Devendra~Singh Chaplot, Diego de~las Casas, Emma~Bou
  Hanna, Florian Bressand, et~al.
\newblock Mixtral of experts.
\newblock \emph{arXiv preprint arXiv:2401.04088}, 2024.

\bibitem[Jiang et~al.(2023{\natexlab{b}})Jiang, Xu, Gao, Sun, Liu, Dwivedi-Yu,
  Yang, Callan, and Neubig]{jiang2023active}
Zhengbao Jiang, Frank Xu, Luyu Gao, Zhiqing Sun, Qian Liu, Jane Dwivedi-Yu,
  Yiming Yang, Jamie Callan, and Graham Neubig.
\newblock Active retrieval augmented generation.
\newblock In Houda Bouamor, Juan Pino, and Kalika Bali (eds.),
  \emph{Proceedings of the 2023 Conference on Empirical Methods in Natural
  Language Processing}, pp.\  7969--7992, Singapore, December
  2023{\natexlab{b}}. Association for Computational Linguistics.
\newblock \doi{10.18653/v1/2023.emnlp-main.495}.
\newblock URL \url{https://aclanthology.org/2023.emnlp-main.495}.

\bibitem[Jiapeng et~al.(2024)Jiapeng, Runze, Yabo, Tong, Mingling, and
  Xiang]{jiapeng2024tree}
Li~Jiapeng, Liu Runze, Li~Yabo, Zhou Tong, Li~Mingling, and Chen Xiang.
\newblock Tree of reviews: A tree-based dynamic iterative retrieval framework
  for multi-hop question answering.
\newblock \emph{arXiv preprint arXiv:2404.14464}, 2024.

\bibitem[Jin \& Han(2011)Jin and Han]{jin2011k}
Xin Jin and Jiawei Han.
\newblock K-means clustering.
\newblock \emph{Encyclopedia of machine learning}, pp.\  563--564, 2011.

\bibitem[Joshi et~al.(2017)Joshi, Choi, Weld, and
  Zettlemoyer]{joshi2017triviaqa}
Mandar Joshi, Eunsol Choi, Daniel~S Weld, and Luke Zettlemoyer.
\newblock Triviaqa: A large scale distantly supervised challenge dataset for
  reading comprehension.
\newblock In \emph{Proceedings of the 55th Annual Meeting of the Association
  for Computational Linguistics (Volume 1: Long Papers)}, pp.\  1601--1611,
  2017.

\bibitem[Kamalloo et~al.(2023)Kamalloo, Dziri, Clarke, and
  Rafiei]{kamalloo2023evaluating}
Ehsan Kamalloo, Nouha Dziri, Charles Clarke, and Davood Rafiei.
\newblock Evaluating open-domain question answering in the era of large
  language models.
\newblock In \emph{Proceedings of the 61st Annual Meeting of the Association
  for Computational Linguistics (Volume 1: Long Papers)}, pp.\  5591--5606,
  2023.

\bibitem[Khandelwal et~al.(2020)Khandelwal, Levy, Jurafsky, Zettlemoyer, and
  Lewis]{khandelwal2020generalization}
Urvashi Khandelwal, Omer Levy, Dan Jurafsky, Luke Zettlemoyer, and Mike Lewis.
\newblock {Generalization through Memorization: Nearest Neighbor Language
  Models}.
\newblock In \emph{International Conference on Learning Representations
  (ICLR)}, 2020.

\bibitem[Kim et~al.(2024)Kim, Nam, Mo, Park, Lee, Seo, Ha, and
  Shin]{kim2024sure}
Jaehyung Kim, Jaehyun Nam, Sangwoo Mo, Jongjin Park, Sang-Woo Lee, Minjoon Seo,
  Jung-Woo Ha, and Jinwoo Shin.
\newblock Sure: Summarizing retrievals using answer candidates for open-domain
  qa of llms.
\newblock \emph{arXiv preprint arXiv:2404.13081}, 2024.

\bibitem[Kojima et~al.(2022)Kojima, Gu, Reid, Matsuo, and
  Iwasawa]{kojima2022large}
Takeshi Kojima, Shixiang~Shane Gu, Machel Reid, Yutaka Matsuo, and Yusuke
  Iwasawa.
\newblock Large language models are zero-shot reasoners.
\newblock In \emph{Advances in Neural Information Processing Systems}, 2022.

\bibitem[Kwon et~al.(2023)Kwon, Li, Zhuang, Sheng, Zheng, Yu, Gonzalez, Zhang,
  and Stoica]{kwon2023efficient}
Woosuk Kwon, Zhuohan Li, Siyuan Zhuang, Ying Sheng, Lianmin Zheng, Cody~Hao Yu,
  Joseph~E. Gonzalez, Hao Zhang, and Ion Stoica.
\newblock Efficient memory management for large language model serving with
  pagedattention.
\newblock In \emph{Proceedings of the ACM SIGOPS 29th Symposium on Operating
  Systems Principles}, 2023.

\bibitem[Leviathan et~al.(2023)Leviathan, Kalman, and
  Matias]{leviathan2023fast}
Yaniv Leviathan, Matan Kalman, and Yossi Matias.
\newblock Fast inference from transformers via speculative decoding.
\newblock In \emph{International Conference on Machine Learning}, pp.\
  19274--19286. PMLR, 2023.

\bibitem[Lewis et~al.(2020)Lewis, Perez, Piktus, Petroni, Karpukhin, Goyal,
  K{\"u}ttler, Lewis, Yih, Rockt{\"a}schel, et~al.]{lewis2020retrieval}
Patrick Lewis, Ethan Perez, Aleksandra Piktus, Fabio Petroni, Vladimir
  Karpukhin, Naman Goyal, Heinrich K{\"u}ttler, Mike Lewis, Wen-tau Yih, Tim
  Rockt{\"a}schel, et~al.
\newblock Retrieval-augmented generation for knowledge-intensive nlp tasks.
\newblock \emph{Advances in Neural Information Processing Systems},
  33:\penalty0 9459--9474, 2020.

\bibitem[Li et~al.(2024)Li, Zhang, Do, Yue, and Chen]{li2024long}
Tianle Li, Ge~Zhang, Quy~Duc Do, Xiang Yue, and Wenhu Chen.
\newblock Long-context llms struggle with long in-context learning.
\newblock \emph{arXiv preprint arXiv:2404.02060}, 2024.

\bibitem[Liu et~al.(2024)Liu, Lin, Hewitt, Paranjape, Bevilacqua, Petroni, and
  Liang]{liu2024lost}
Nelson~F Liu, Kevin Lin, John Hewitt, Ashwin Paranjape, Michele Bevilacqua,
  Fabio Petroni, and Percy Liang.
\newblock Lost in the middle: How language models use long contexts.
\newblock \emph{Transactions of the Association for Computational Linguistics},
  12, 2024.

\bibitem[Luo et~al.(2023{\natexlab{a}})Luo, Chuang, Gong, Zhang, Kim, Wu, Fox,
  Meng, and Glass]{luo2023sail}
Hongyin Luo, Yung-Sung Chuang, Yuan Gong, Tianhua Zhang, Yoon Kim, Xixin Wu,
  Danny Fox, Helen Meng, and James Glass.
\newblock Sail: Search-augmented instruction learning.
\newblock \emph{arXiv preprint arXiv:2305.15225}, 2023{\natexlab{a}}.

\bibitem[Luo et~al.(2023{\natexlab{b}})Luo, Yang, Meng, Li, Zhou, and
  Zhang]{luo2023empirical}
Yun Luo, Zhen Yang, Fandong Meng, Yafu Li, Jie Zhou, and Yue Zhang.
\newblock An empirical study of catastrophic forgetting in large language
  models during continual fine-tuning.
\newblock \emph{arXiv preprint arXiv:2308.08747}, 2023{\natexlab{b}}.

\bibitem[Ma et~al.(2023)Ma, Gong, He, Zhao, and Duan]{ma2023query}
Xinbei Ma, Yeyun Gong, Pengcheng He, Hai Zhao, and Nan Duan.
\newblock Query rewriting in retrieval-augmented large language models.
\newblock In \emph{Proceedings of the 2023 Conference on Empirical Methods in
  Natural Language Processing}, pp.\  5303--5315, 2023.

\bibitem[Ma et~al.(2024)Ma, Yang, Xiong, Chen, Yu, Zhang, May, Zettlemoyer,
  Levy, and Zhou]{ma2024megalodon}
Xuezhe Ma, Xiaomeng Yang, Wenhan Xiong, Beidi Chen, Lili Yu, Hao Zhang,
  Jonathan May, Luke Zettlemoyer, Omer Levy, and Chunting Zhou.
\newblock Megalodon: Efficient llm pretraining and inference with unlimited
  context length.
\newblock \emph{arXiv preprint arXiv:2404.08801}, 2024.

\bibitem[Mallen et~al.(2023)Mallen, Asai, Zhong, Das, Khashabi, and
  Hajishirzi]{mallen2023not}
Alex Mallen, Akari Asai, Victor Zhong, Rajarshi Das, Daniel Khashabi, and
  Hannaneh Hajishirzi.
\newblock When not to trust language models: Investigating effectiveness of
  parametric and non-parametric memories.
\newblock In \emph{Proceedings of the 61st Annual Meeting of the Association
  for Computational Linguistics (Volume 1: Long Papers)}, pp.\  9802--9822,
  2023.

\bibitem[Miao et~al.(2024)Miao, Oliaro, Zhang, Cheng, Wang, Zhang, Wong, Zhu,
  Yang, Shi, et~al.]{miao2024specinfer}
Xupeng Miao, Gabriele Oliaro, Zhihao Zhang, Xinhao Cheng, Zeyu Wang, Zhengxin
  Zhang, Rae Ying~Yee Wong, Alan Zhu, Lijie Yang, Xiaoxiang Shi, et~al.
\newblock Specinfer: Accelerating large language model serving with tree-based
  speculative inference and verification.
\newblock In \emph{Proceedings of the 29th ACM International Conference on
  Architectural Support for Programming Languages and Operating Systems, Volume
  3}, pp.\  932--949, 2024.

\bibitem[Mihaylov et~al.(2018)Mihaylov, Clark, Khot, and
  Sabharwal]{mihaylov2018can}
Todor Mihaylov, Peter Clark, Tushar Khot, and Ashish Sabharwal.
\newblock Can a suit of armor conduct electricity? a new dataset for open book
  question answering.
\newblock In \emph{Proceedings of the 2018 Conference on Empirical Methods in
  Natural Language Processing}, pp.\  2381--2391, 2018.

\bibitem[Min et~al.(2023)Min, Krishna, Lyu, Lewis, Yih, Koh, Iyyer,
  Zettlemoyer, and Hajishirzi]{min2023factscore}
Sewon Min, Kalpesh Krishna, Xinxi Lyu, Mike Lewis, Wen-tau Yih, Pang Koh, Mohit
  Iyyer, Luke Zettlemoyer, and Hannaneh Hajishirzi.
\newblock Factscore: Fine-grained atomic evaluation of factual precision in
  long form text generation.
\newblock In \emph{Proceedings of the 2023 Conference on Empirical Methods in
  Natural Language Processing}, pp.\  12076--12100, 2023.

\bibitem[Peng et~al.(2024)Peng, Zhang, Wang, Srinivasa, Liu, Wang, and
  Shang]{peng2024answer}
Letian Peng, Yuwei Zhang, Zilong Wang, Jayanth Srinivasa, Gaowen Liu, Zihan
  Wang, and Jingbo Shang.
\newblock Answer is all you need: Instruction-following text embedding via
  answering the question.
\newblock \emph{arXiv preprint arXiv:2402.09642}, 2024.

\bibitem[Petroni et~al.(2021)Petroni, Piktus, Fan, Lewis, Yazdani, De~Cao,
  Thorne, Jernite, Karpukhin, Maillard, et~al.]{petroni2021kilt}
Fabio Petroni, Aleksandra Piktus, Angela Fan, Patrick Lewis, Majid Yazdani,
  Nicola De~Cao, James Thorne, Yacine Jernite, Vladimir Karpukhin, Jean
  Maillard, et~al.
\newblock Kilt: a benchmark for knowledge intensive language tasks.
\newblock In \emph{Proceedings of the 2021 Conference of the North American
  Chapter of the Association for Computational Linguistics: Human Language
  Technologies}, pp.\  2523--2544, 2021.

\bibitem[Rasley et~al.(2020)Rasley, Rajbhandari, Ruwase, and
  He]{rasley2020deepspeed}
Jeff Rasley, Samyam Rajbhandari, Olatunji Ruwase, and Yuxiong He.
\newblock Deepspeed: System optimizations enable training deep learning models
  with over 100 billion parameters.
\newblock In \emph{Proceedings of the 26th ACM SIGKDD International Conference
  on Knowledge Discovery \& Data Mining}, pp.\  3505--3506, 2020.

\bibitem[Reid et~al.(2024)Reid, Savinov, Teplyashin, Lepikhin, Lillicrap,
  Alayrac, Soricut, Lazaridou, Firat, Schrittwieser, et~al.]{reid2024gemini}
Machel Reid, Nikolay Savinov, Denis Teplyashin, Dmitry Lepikhin, Timothy
  Lillicrap, Jean-baptiste Alayrac, Radu Soricut, Angeliki Lazaridou, Orhan
  Firat, Julian Schrittwieser, et~al.
\newblock Gemini 1.5: Unlocking multimodal understanding across millions of
  tokens of context.
\newblock \emph{arXiv preprint arXiv:2403.05530}, 2024.

\bibitem[Sarthi et~al.(2024)Sarthi, Abdullah, Tuli, Khanna, Goldie, and
  Manning]{sarthi2024raptor}
Parth Sarthi, Salman Abdullah, Aditi Tuli, Shubh Khanna, Anna Goldie, and
  Christopher~D Manning.
\newblock Raptor: Recursive abstractive processing for tree-organized
  retrieval.
\newblock \emph{arXiv preprint arXiv:2401.18059}, 2024.

\bibitem[Schick et~al.(2024)Schick, Dwivedi-Yu, Dess{\`\i}, Raileanu, Lomeli,
  Hambro, Zettlemoyer, Cancedda, and Scialom]{schick2024toolformer}
Timo Schick, Jane Dwivedi-Yu, Roberto Dess{\`\i}, Roberta Raileanu, Maria
  Lomeli, Eric Hambro, Luke Zettlemoyer, Nicola Cancedda, and Thomas Scialom.
\newblock Toolformer: Language models can teach themselves to use tools.
\newblock \emph{Advances in Neural Information Processing Systems}, 36, 2024.

\bibitem[Shi et~al.(2024)Shi, Zhang, Sun, Gao, Ren, Chen, and
  Ren]{shi2024generate}
Zhengliang Shi, Shuo Zhang, Weiwei Sun, Shen Gao, Pengjie Ren, Zhumin Chen, and
  Zhaochun Ren.
\newblock Generate-then-ground in retrieval-augmented generation for multi-hop
  question answering.
\newblock \emph{arXiv preprint arXiv:2406.14891}, 2024.

\bibitem[Stelmakh et~al.(2022)Stelmakh, Luan, Dhingra, and
  Chang]{stelmakh2022asqa}
Ivan Stelmakh, Yi~Luan, Bhuwan Dhingra, and Ming-Wei Chang.
\newblock Asqa: Factoid questions meet long-form answers.
\newblock In \emph{Proceedings of the 2022 Conference on Empirical Methods in
  Natural Language Processing}, pp.\  8273--8288, 2022.

\bibitem[Stern et~al.(2018)Stern, Shazeer, and Uszkoreit]{stern2018blockwise}
Mitchell Stern, Noam Shazeer, and Jakob Uszkoreit.
\newblock Blockwise parallel decoding for deep autoregressive models.
\newblock \emph{Advances in Neural Information Processing Systems}, 31, 2018.

\bibitem[Team et~al.(2023)Team, Anil, Borgeaud, Wu, Alayrac, Yu, Soricut,
  Schalkwyk, Dai, Hauth, et~al.]{team2023gemini}
Gemini Team, Rohan Anil, Sebastian Borgeaud, Yonghui Wu, Jean-Baptiste Alayrac,
  Jiahui Yu, Radu Soricut, Johan Schalkwyk, Andrew~M Dai, Anja Hauth, et~al.
\newblock Gemini: a family of highly capable multimodal models.
\newblock \emph{arXiv preprint arXiv:2312.11805}, 2023.

\bibitem[Team et~al.(2024)Team, Riviere, Pathak, Sessa, Hardin, Bhupatiraju,
  Hussenot, Mesnard, Shahriari, Ram{\'e}, et~al.]{team2024gemma}
Gemma Team, Morgane Riviere, Shreya Pathak, Pier~Giuseppe Sessa, Cassidy
  Hardin, Surya Bhupatiraju, L{\'e}onard Hussenot, Thomas Mesnard, Bobak
  Shahriari, Alexandre Ram{\'e}, et~al.
\newblock Gemma 2: Improving open language models at a practical size.
\newblock \emph{arXiv preprint arXiv:2408.00118}, 2024.

\bibitem[Trivedi et~al.(2022)Trivedi, Balasubramanian, Khot, and
  Sabharwal]{trivedi2022musique}
Harsh Trivedi, Niranjan Balasubramanian, Tushar Khot, and Ashish Sabharwal.
\newblock Musique: Multihop questions via single-hop question composition.
\newblock \emph{Transactions of the Association for Computational Linguistics},
  10:\penalty0 539--554, 2022.

\bibitem[Wang et~al.(2023{\natexlab{a}})Wang, Ivison, Dasigi, Hessel, Khot,
  Chandu, Wadden, MacMillan, Smith, Beltagy, et~al.]{wang2023far}
Yizhong Wang, Hamish Ivison, Pradeep Dasigi, Jack Hessel, Tushar Khot,
  Khyathi~Raghavi Chandu, David Wadden, Kelsey MacMillan, Noah~A Smith,
  Iz~Beltagy, et~al.
\newblock How far can camels go.
\newblock \emph{Exploring the state of instruction tuning on open resources},
  2023{\natexlab{a}}.

\bibitem[Wang et~al.(2023{\natexlab{b}})Wang, Araki, Jiang, Parvez, and
  Neubig]{wang2023learning}
Zhiruo Wang, Jun Araki, Zhengbao Jiang, Md~Rizwan Parvez, and Graham Neubig.
\newblock Learning to filter context for retrieval-augmented generation.
\newblock \emph{arXiv preprint arXiv:2311.08377}, 2023{\natexlab{b}}.

\bibitem[Wang et~al.(2024)Wang, Liu, Lin, Li, Ma, and Liang]{wang2024rat}
Zihao Wang, Anji Liu, Haowei Lin, Jiaqi Li, Xiaojian Ma, and Yitao Liang.
\newblock Rat: Retrieval augmented thoughts elicit context-aware reasoning in
  long-horizon generation.
\newblock \emph{arXiv preprint arXiv:2403.05313}, 2024.

\bibitem[Wolf et~al.(2019)Wolf, Debut, Sanh, Chaumond, Delangue, Moi, Cistac,
  Rault, Louf, Funtowicz, et~al.]{wolf2019huggingface}
Thomas Wolf, Lysandre Debut, Victor Sanh, Julien Chaumond, Clement Delangue,
  Anthony Moi, Pierric Cistac, Tim Rault, R{\'e}mi Louf, Morgan Funtowicz,
  et~al.
\newblock Huggingface's transformers: State-of-the-art natural language
  processing.
\newblock \emph{arXiv preprint arXiv:1910.03771}, 2019.

\bibitem[Xia et~al.(2023)Xia, Ge, Wang, Chen, Wei, and Sui]{xia2023speculative}
Heming Xia, Tao Ge, Peiyi Wang, Si-Qing Chen, Furu Wei, and Zhifang Sui.
\newblock Speculative decoding: Exploiting speculative execution for
  accelerating seq2seq generation.
\newblock In \emph{Findings of the Association for Computational Linguistics:
  EMNLP 2023}, pp.\  3909--3925, 2023.

\bibitem[Xia et~al.(2024{\natexlab{a}})Xia, Yang, Dong, Wang, Li, Ge, Liu, Li,
  and Sui]{xia2024unlocking}
Heming Xia, Zhe Yang, Qingxiu Dong, Peiyi Wang, Yongqi Li, Tao Ge, Tianyu Liu,
  Wenjie Li, and Zhifang Sui.
\newblock Unlocking efficiency in large language model inference: A
  comprehensive survey of speculative decoding.
\newblock \emph{arXiv preprint arXiv:2401.07851}, 2024{\natexlab{a}}.

\bibitem[Xia et~al.(2024{\natexlab{b}})Xia, Wang, Liang, Zhang, Zhou, Deng, Yu,
  and Xiao]{xia2024ground}
Sirui Xia, Xintao Wang, Jiaqing Liang, Yifei Zhang, Weikang Zhou, Jiaji Deng,
  Fei Yu, and Yanghua Xiao.
\newblock Ground every sentence: Improving retrieval-augmented llms with
  interleaved reference-claim generation.
\newblock \emph{arXiv preprint arXiv:2407.01796}, 2024{\natexlab{b}}.

\bibitem[Xie et~al.(2023)Xie, Zhang, Chen, Lou, and Su]{xie2023adaptive}
Jian Xie, Kai Zhang, Jiangjie Chen, Renze Lou, and Yu~Su.
\newblock Adaptive chameleon or stubborn sloth: Revealing the behavior of large
  language models in knowledge conflicts.
\newblock In \emph{The Twelfth International Conference on Learning
  Representations}, 2023.

\bibitem[Xu et~al.(2023)Xu, Shi, and Choi]{xu2023recomp}
Fangyuan Xu, Weijia Shi, and Eunsol Choi.
\newblock Recomp: Improving retrieval-augmented lms with compression and
  selective augmentation.
\newblock \emph{arXiv preprint arXiv:2310.04408}, 2023.

\bibitem[Yan et~al.(2024)Yan, Gu, Zhu, and Ling]{yan2024corrective}
Shi-Qi Yan, Jia-Chen Gu, Yun Zhu, and Zhen-Hua Ling.
\newblock Corrective retrieval augmented generation.
\newblock \emph{arXiv preprint arXiv:2401.15884}, 2024.

\bibitem[Yang et~al.(2018)Yang, Qi, Zhang, Bengio, Cohen, Salakhutdinov, and
  Manning]{yang2018hotpotqa}
Zhilin Yang, Peng Qi, Saizheng Zhang, Yoshua Bengio, William Cohen, Ruslan
  Salakhutdinov, and Christopher~D Manning.
\newblock Hotpotqa: A dataset for diverse, explainable multi-hop question
  answering.
\newblock In \emph{Proceedings of the 2018 Conference on Empirical Methods in
  Natural Language Processing}, pp.\  2369--2380, 2018.

\bibitem[Yoran et~al.(2023)Yoran, Wolfson, Ram, and Berant]{yoran2023making}
Ori Yoran, Tomer Wolfson, Ori Ram, and Jonathan Berant.
\newblock Making retrieval-augmented language models robust to irrelevant
  context.
\newblock \emph{arXiv preprint arXiv:2310.01558}, 2023.

\bibitem[Yu et~al.(2023)Yu, Zhang, Pan, Ma, Wang, and Yu]{yu2023chain}
Wenhao Yu, Hongming Zhang, Xiaoman Pan, Kaixin Ma, Hongwei Wang, and Dong Yu.
\newblock Chain-of-note: Enhancing robustness in retrieval-augmented language
  models.
\newblock \emph{arXiv preprint arXiv:2311.09210}, 2023.

\bibitem[Zhang et~al.(2023{\natexlab{a}})Zhang, Wang, Li, Shou, Chen, Chen, and
  Mehrotra]{zhang2023draft}
Jun Zhang, Jue Wang, Huan Li, Lidan Shou, Ke~Chen, Gang Chen, and Sharad
  Mehrotra.
\newblock Draft \& verify: Lossless large language model acceleration via
  self-speculative decoding.
\newblock \emph{arXiv preprint arXiv:2309.08168}, 2023{\natexlab{a}}.

\bibitem[Zhang et~al.(2023{\natexlab{b}})Zhang, Luo, Chuang, Fang, Gaitskell,
  Hartvigsen, Wu, Fox, Meng, and Glass]{zhang2023interpretable}
Tianhua Zhang, Hongyin Luo, Yung-Sung Chuang, Wei Fang, Luc Gaitskell, Thomas
  Hartvigsen, Xixin Wu, Danny Fox, Helen Meng, and James Glass.
\newblock Interpretable unified language checking.
\newblock \emph{arXiv preprint arXiv:2304.03728}, 2023{\natexlab{b}}.

\end{thebibliography}
\bibliographystyle{iclr2025_conference}


\newpage
\appendix
\section*{Appendix}

\section{Instruction-Tuning Settings}\label{appendix:setting}
We construct our training dataset for the RAG drafter from diverse instruction-following pairs. We sample instances from Open-Instruct processed data~\citep{wang2023far} and knowledge-intensive datasets~\citep{petroni2021kilt,stelmakh2022asqa,mihaylov2018can}. We augment the instruction-following pairs with retrieved documents and generated rationale. We use the off-the-shelf dense retriever Contriever-MS MARCO~\citep{izacard2021unsupervised} to retrieve up to 10 documents for each pair and use Gemini-Ultra~\citep{team2023gemini} to generate rationale. In total, we acquire a dataset of 40k instances. We use Mistral$_\text{7B}$ (v0.1) as our base LM for the RAG drafter. We reproduce the performance of Self-RAG~\citep{asai2023self} and CRAG~\citep{yan2024corrective} with Mistral$_\text{7B}$ (v0.1) for a fair comparison. We implement the training scripts using the Transformers library from Hugging Face \citep{wolf2019huggingface}. We employ DeepSpeed \citep{rasley2020deepspeed} to accelerate the training process. All experiments are conducted on a Linux server equipped with 16 Nvidia A100-SXM4-40GB GPUs.


Additionally, we replace Gemini-Ultra~\citep{team2023gemini} with GPT-4o~\citep{achiam2023gpt} when curating the instruction-tuning data for our RAG drafter to investigate the effects of different LLMs. These results demonstrate that \method maintains its performance advantage, even when trained on data curated by GPT-4o. It consistently outperforms the baselines from Standard RAG, SelfRAG, and CRAG, further validating the effectiveness of our approach.

\begin{table}[htbp]
  \centering
  \caption{RAG results on TriviaQA, PubHealth, ARC-Challenge with the RAG drafter trained on instruction-tuning data curated by GPT-4o and Gemini-Ultra.}
  \small
    \begin{tabular}{lccc}
    \toprule
          & \textbf{Trivial QA} & \textbf{PubHealth} & \textbf{ARC-C} \\
    \midrule
    $\mathcal{M}_\text{Verifier-7B}$ + $\mathcal{M}^{\text{Gemini-U}}_\text{Drafter-7B}$ & 73.91 & 75.79 & 76.19 \\
    $\mathcal{M}_\text{Verifier-8x7B}$ + $\mathcal{M}^{\text{Gemini-U}}_\text{Drafter-7B}$ & 74.24 & 76.60 &  80.55 \\
    \midrule
    $\mathcal{M}_\text{Verifier-7B}$ + $\mathcal{M}^{\text{GPT-4o}}_\text{Drafter-7B}$ & 72.24 & 73.05 & 76.54 \\
    $\mathcal{M}_\text{Verifier-8x7B}$ + $\mathcal{M}^{\text{GPT-4o}}_\text{Drafter-7B}$  & 73.58 & 73.35 & 80.63 \\
    \bottomrule
    \end{tabular}%
  \label{tab:gpt4}%
\end{table}%

Moreover, we use an instruction-tuned Gemma-2$_\text{2B}$~\citep{team2024gemma} as the RAG drafter and the frozen Mistral$_\text{7B}$ or Mixtral$_\text{8x7B}$ as the RAG verifier. We report the performance analysis in Table~\ref{tab:gemma2}. These results suggest that Gemma-2$_\text{2B}$ provides a promising avenue for future work of further optimization.

\begin{table}[htbp]
  \centering
  \caption{RAG results on TriviaQA, PubHealth, ARC-Challenge with an instruction-tuned Gemma-2$_\text{2B}$ or Mistral$_\text{7B}$ as the RAG drafter and a Mixtral$_\text{8x7B}$ as the RAG verifier.}
  \small
    \begin{tabular}{lccc}
    \toprule
          & \textbf{Trivial QA} & \textbf{PubHealth} & \textbf{ARC-C} \\
    \midrule
    $\mathcal{M}_\text{Verifier-8x7B}$ + $\mathcal{M}_\text{Drafter-2B}$ & 67.22 & 72.14 & 67.92 \\
    $\mathcal{M}_\text{Verifier-8x7B}$ + $\mathcal{M}_\text{Drafter-7B}$ & 74.24 & 76.60 &  80.55 \\
    \bottomrule
    \end{tabular}%
  \label{tab:gemma2}%
\end{table}%

\section{Effects of Instruction Tuning}\label{appendix:m1kn}

In \method, we introduce a framework that combines the RAG drafter and the verifier. In this ablation study, we directly feed all documents to the RAG drafter and generate one draft ($m=1$, $k=\text{total \# of documents}$). As shown in Table~\ref{tab:m1kn}, we observe that instruction tuning effectively enhances the document understanding capability of the RAG drafter, as it outperforms both Mistral$_\text{7B}$ and Mistral-Instruct$_\text{7B}$. However, there remains a gap compared to \method, showing the effectiveness of the drafting and verification framework.

\begin{table}[htbp]
  \centering
  \vspace{-3mm}
  \caption{RAG results on TriviaQA and PubHealth ($m=1$, $k=\text{total \# of docs}$)}
  \small
    \begin{tabular}{lcc}
    \toprule
     & \textbf{TrivialQA} & \textbf{PubHealth} \\
    \midrule
    Mistral$_\text{7B}$ & 54.15 & 34.85 \\
    Mistral-Instruct$_\text{7B}$ & 67.11 & 42.15 \\
    \midrule
    $\mathcal{M}_\text{Drafter-7B}$ ($m=1$, $k=10$) & {73.23} & {65.25} \\
    \midrule
    $\mathcal{M}_\text{Verifier-7B}$ + $\mathcal{M}_\text{Drafter-7B}$ ($m=5$, $k=2$) & {\textbf{73.91}} & {\textbf{75.79}} \\
    \bottomrule
    \end{tabular}%
  \label{tab:m1kn}%
\end{table}%

\section{Effects of Self-Reflection Statement}

We use ``Do you think the explanation supports the answers? (Yes or No)'' as the self-reflection statement in our main results. In this study, we replace it with other alternatives to see how the self-reflection statement affects the accuracy. The results are reported in Table~\ref{tab:reflection-statement}. We observe that the performance does not change a lot given different self-reflection statements, which shows the stable verification capability of the generalist LMs by language modeling objective.

\begin{table}[htbp]
  \centering
  \caption{Performance analysis of \method with different self-reflection statements $R$ when computing the self-reflection score $\rho_\text{Self-reflect}=P(\texttt{"Yes"}|Q,\alpha,\beta,R)$, where $Q$ is the query, $\alpha,\beta$ are the generated answer draft and rationale.}
  \small
  \resizebox{\linewidth}{!}{
    \setlength{\tabcolsep}{4mm}{
    \begin{tabular}{lcc}
    \toprule
    \textbf{Reflection Statement} & \textbf{TriviaQA} & \textbf{PubHealth} \\
    \midrule
    \textit{Do you think the explanation supports the answers? (Yes or No)} & 74.24 & 76.60 \\
    \textit{Does the rationale support the answer? (Yes or No)} & 74.22 & 76.09\\
    \textit{What do you think about the rationale? A good one? (Yes or No)} & 74.25 & 75.79  \\
    \textit{Is the rationale good enough to support the answer? (Yes or No)} & 74.39 & 76.29 \\
    \bottomrule
    \end{tabular}%
    }
    }
  \label{tab:reflection-statement}%
\end{table}%

\section{Effects of Rationale Generation}\label{appendix:rationale-generation}

We acknowledge that the generation of rationale potentially increases the inference cost during the drafting stage while this is crucial for the verifier in our method to assess the quality and reliability of generated drafts. And, the potential overhead can be mitigated through efficient parallel inference.

To further study the impact of rationale generation, we finetune the RAG drafter without rationale. We denote this setting as: \textit{without rationale in drafting}. Similarly, \textit{with rationale/doc in verification} indicates that we use the generated rationale or the retrieved documents as reference during the verification stage. We use $\mathcal{M}_\text{Verifier-8x7B}$ + $\mathcal{M}_\text{Drafter-7B}$ as a running example. The results are shown in Table~\ref{tab:rationale-ablation-2}.

\begin{table}[htbp]
  \centering
  \caption{Ablation study on the draft generation in the drafting and verification stages on TriviaQA and PubHealth.}
  \small
    \begin{tabular}{llccc}
    \toprule
    \textbf{Drafting} & \textbf{Verification} & \textbf{TrivialQA} & \textbf{PubHealth} & \textbf{ARC-C} \\
    \midrule
    \textit{w/o rationale} & \textit{w/ doc} & 70.86 & 75.28 & 78.67 \\
    \textit{w/ rationale} & \textit{w/ doc} & 74.08 & 76.09 & 80.46 \\
    \textit{w/ rationale} & \textit{w/ rationale} & 74.24 & 76.60  & 80.55 \\
    \bottomrule
    \end{tabular}%
  \label{tab:rationale-ablation-2}%
\end{table}%

\paragraph{Better answer drafting} As explored in \citet{hsieh2023distilling}, incorporating rationale generation during instruction-tuning can lead to the RAG drafter producing higher-quality answer drafts. The results in Table~\ref{tab:rationale-ablation-2} clearly demonstrate this. We observe a significant performance drop across all three benchmarks when the RAG drafter is finetuned without the rationale component,

\paragraph{Lower latency and cost in verification} We verify each draft against the rationale instead of the retrieved documents. From the ablation results, these generated rationales serve as high-quality grounding facts, improving verification performance compared to using the retrieved documents.

\section{Effects of Different Volume of Training Data}

We acknowledge the importance of evaluating our framework's performance across different training data volumes. Our primary experiment utilized 40,059 instances to train our drafter model. To thoroughly assess scaling effects, we conducted additional experiments using incremental subsets of 10,000, 20,000, and 30,000 training instances. The results of these systematic evaluations are detailed in Table~\ref{tab:diff-volumn}.

\begin{table}[ht]
  \centering
  \caption{Performance analysis of \method with different volumn of instruction-tuning data.}
  \small
    \begin{tabular}{lcc}
    \toprule
    $\mathcal{M}_\text{Verifier-8x7B}$ + $\mathcal{M}_\text{Drafter-7B}$ & \textbf{TriviaQA} & \textbf{PubHealth} \\
    \midrule
    10,000  & 71.69 & 72.34 \\
    20,000  & 72.64 & 72.44 \\
    30,000  & 73.20 & 74.37 \\
    \midrule
    \textbf{Total (40,059)} & \textbf{74.24} & \textbf{76.60} \\
    \bottomrule
    \end{tabular}%
  \label{tab:diff-volumn}%
\end{table}%

From the table, we can conclude that increasing the volume of training data leads to improved performance. Specifically, the model's accuracy continues to rise as more instances are included, with the highest performance observed at 40,059 instances. This suggests that larger training datasets contribute positively to the performance of our drafter-verifier framework, indicating that scaling up data size could enhance the robustness of the model.

\section{Efficacy of \method in Multi-hop Reasoning}\label{sec:multihop}

We further validate \method in the scenario of multi-hop reasoning. One of the key challenges of multi-hop reasoning is to effectively combine multiple pieces of evidence to arrive at the final answer. Indeed, the ability to verify or contrast information across documents is crucial to solve complex questions. We compare the performance of \method with baselines on MuSiQue~\citep{trivedi2022musique} and HotpotQA~\citep{yang2018hotpotqa}, two multi-hop reasoning benchmarks. We randomly sample 500 examples from the validation set of HotpotQA as the test set in our experiment. We adopt the same setting as MuSiQue on HotpotQA. The results are in Table~\ref{tab:multihop-result}. We find that our \method achieves the best performance. Specifically, \method improves accuracy by 2.15\% on MuSiQue and by a substantial 5.4\% on HotpotQA.
\begin{table}[htbp]
  \centering
  \caption{RAG results on MuSiQue and HotpotQA}
    \begin{tabular}{lcc}
    \toprule
          & \textbf{MuSiQue} & \textbf{HotpotQA} \\
    \midrule
    Mixtral-Instruct$_\text{8x7B}$ & 29.42 & 43.60 \\
    Self-RAG$_\text{Mistral-7B}$ & 21.72 & 27.20 \\
    \midrule
    $\mathcal{M}_\text{Verifier-7B}$ + $\mathcal{M}_\text{Drafter-7B}$& 31.03 & 47.60 \\
    $\mathcal{M}_\text{Verifier-8x7B}$ + $\mathcal{M}_\text{Drafter-7B}$ & \textbf{31.57} & \textbf{49.00} \\
    \bottomrule
    \end{tabular}%
  \label{tab:multihop-result}%
\end{table}%

Our approach tackles this challenge by multi-perspective sampling when selecting documents for each draft (Section~\ref{sec:drafter}). We cluster the retrieved documents into distinct topics using an instruction-aware embedding model~\citep{peng2024answer}. Then, we sample one document from each cluster to form a diverse document subset, ensuring each drafter receives a variety of perspectives from the retrieval results. To validate the efficacy of this strategy, we further conduct an ablation study on MuSiQue and HotpotQA in Table~\ref{tab:multihop}. From the table, our sampling strategy effectively guarantees the diversity of information within the supporting document subsets, leading to improved performance of \method on these tasks.

\begin{table}[ht]
  \centering
  \caption{Ablation study of multi-perspective sampling on multi-hop reasoning benchmarks: MuSiQue, HotpotQA.}
    \begin{tabular}{lcc}
    \toprule
          & \textbf{MuSiQue} & \textbf{HotpotQA} \\
    \midrule
    Random sampling & 29.33 & 48.2 \\
    Multi-perspective sampling & \textbf{31.57} & \textbf{49.00} \\
    \bottomrule
    \end{tabular}%
  \label{tab:multihop}%
\end{table}%

\subsection{Performance Breakdown on HotpotQA}

HotpotQA includes two types of quesitions: bridge-type questions in HotpotQA require a two-step reasoning process where the answer to the first step is crucial for answering the second. For example:
\begin{itemize}
    \item \textit{"When was the singer and songwriter of Radiohead born?"}
    \begin{itemize}
        \item Step 1: Who is the singer and songwriter of Radiohead? $\to$ Thom Yorke
        \item Step 2: When was [Thom Yorke](answer of step 1) born? $\to$ October 7, 1968
        \item Final answer: October 7, 1968
    \end{itemize}
\end{itemize}

In contrast, comparison-type questions also involve two steps, but the answers to each step are independent of each other. For example:
\begin{itemize}
    \item \textit{"Who was born first, Morgan Llywelyn or Robert Jordan?"}
    \begin{itemize}
        \item Step 1: What’s Morgan Llywelyn’s DOB? $\to$ December 3, 1937
        \item Step 2: What’s Robert Jordan’s DOB? $\to$ October 17, 1948
        \item Final answer: Morgan Llywelyn
    \end{itemize}
\end{itemize}

\begin{table}[h!]
\centering
\caption{Performance of \method for different question types}
\begin{tabular}{lcc}
\toprule
\textbf{Question Type} & \textbf{\# of Questions} & \method \\ 
\midrule
Bridge-type            & 400                      & 41.75                               \\ 
Comparison-type        & 100                      & 78.00                               \\ 
\midrule
Overall                & 500                      & 49.00                               \\ 
\bottomrule
\end{tabular}
  \label{tab:hotpot-breakdown}%
\end{table}

We report the performance breakdown of \method on HotpotQA in Table~\ref{tab:hotpot-breakdown}. The results demonstrate a superior performance on comparison-type questions with multi-perspective sampling. This aligns with our expectations, as multi-perspective sampling ensures the document subset covers the diverse topics necessary for answering comparison-type questions. Revisiting the example above, \textit{"Who was born first, Morgan Llywelyn or Robert Jordan?"}, with $k=4$, our approach clusters retrieved documents into four groups. Group 0 and 3 focus on Morgan, while group 1 and 2 focus on Robert. As we sample one document from each group for the drafters, this clustering result ensures each drafter receives documents about both individuals. This balanced information distribution is crucial for the comparison-type questions. In contrast, random sampling risks providing a drafter with information about only one person, yielding a suboptimal draft.

On the other hand, we also observe that the multi-perspective sampling is less helpful for bridge-type questions. These questions require the LLM to first identify the \textit{"bridge entity"} (e.g., Thom Yorke in the earlier example), a task our current work isn't explicitly designed for. While multi-perspective sampling effectively covers different topics in the drafts and the map-reduce approach accelerates inference, they might not directly contribute to pinpointing the "bridge entity" - the key to answering bridge-type questions.

We believe our framework could be effectively combined with other techniques specifically designed for bridge-type questions, such as those proposed in \citet{xia2024ground,jiapeng2024tree,shi2024generate}. For instance, the Tree-of-Reviews (ToR) framework, introduced in \citet{jiapeng2024tree}, addresses multi-hop reasoning problems by dynamically initiating new searches based on previously retrieved documents and constructing various reasoning paths. This dynamic searching strategy can be integrated into our \method, enabling each drafter to answer bridge-type questions more effectively.

\newpage

\section{Prompt of Rationale Generation}\label{appendix:gemini-prompt}
\begin{figure*}[ht]
\begin{tcolorbox}[left=1mm,right=1mm,top=0.mm, bottom=0mm,colback=white]
\begin{lstlisting}[style=demo]
========================================= Prompt ========================================= 

# Memorize this piece of evidence in mind and use it as if you already know it.
# Evidence: State religion
Despite enjoying considerable popular support, Christianity was still not the official state religion in Rome, although it was in some neighboring states such as Armenia, Iberia, and Aksum. Roman Religion (Neoplatonic Hellenism) was restored for a time by the Emperor Julian from 361 to 363. Julian does not appear to have reinstated the persecutions of the earlier Roman emperors. Catholic Christianity, as opposed to Arianism and other ideologies deemed heretical, was declared to be the state religion of the Roman Empire on 27 February 380 by the decree "De fide catolica".

# Instruction: When did christianity become official religion of Rome?

# Response: 380 AD

# Rationale: (complete the following sentence with details from the evidence; you can only use the information from the evidence) 

======================================= Completion =======================================

Catholic Christianity was declared the official state religion of the Roman Empire on February 27th, 380, through the decree "De fide catolica". 
\end{lstlisting}
\end{tcolorbox}
\caption{Prompt of Rationale Generation for Gemini-Ultra}
\label{fig:gemini-prompt}
\end{figure*}

\section{Prompt of RAG Drafting}\label{appendix:rag-prompt}
\begin{figure*}[ht]
\begin{tcolorbox}[left=1mm,right=1mm,top=0.mm, bottom=0mm,colback=white]
\begin{lstlisting}[style=demo]
========================================= Prompt ========================================= 

Response to the instruction. Also provide rationale for your response.
## Instruction: In Buddhism, what is the state of blissful repose or absolute existence by someone relieved of the necessity of rebirth?
## Evidence: 
[1] Buddhism
Nirvana literally means "blowing out, quenching, becoming extinguished". In early Buddhist texts, it is the state of restraint and self-control that leads to the "blowing out" and the ending of the cycles of sufferings associated with rebirths and redeaths. Many later Buddhist texts describe nirvana as identical with "anatta" with complete "emptiness, nothingness". In some texts, the state is described with greater detail, such as passing through the gate of emptiness ("sunyata") realizing that there"
[2] Salvation
It includes a variety of disciplines, such as yoga and meditation. Nirvana is the profound peace of mind that is acquired with moksha (liberation). In Buddhism and Jainism, it is the state of being free from suffering. In Hindu philosophy, it is union with the Brahman (Supreme Being). The word literally means "blown out" (as in a candle) and refers, in the Buddhist context, to the blowing out of the fires of desire, aversion, and delusion, and the imperturbable stillness of mind acquired thereafter. In Theravada Buddhism the emphasis is on one's

======================================= Completion =======================================
## Rationale: Nirvana literally means 'blowing out, quenching, becoming extinguished'. It is described as a state of "restraint and self-control" that leads to the "blowing out" and the ending of the cycles of sufferings associated with rebirths and redeaths.
## Response: In Buddhism, the state of blissful repose or absolute existence by someone relieved of the necessity of rebirth is called Nirvana.

\end{lstlisting}
\end{tcolorbox}
\caption{Prompt of RAG Drafting}
\label{fig:rag-prompt}
\end{figure*}


\newpage

\section{Prompt of Standard RAG}\label{appendix:baseline-prompt}
\begin{figure*}[ht]
\begin{tcolorbox}[left=1mm,right=1mm,top=0.mm, bottom=0mm,colback=white]
\begin{lstlisting}[style=demo]
========================================= Prompt ========================================= 

Below is an instruction that describes a task. Write a response that appropriately completes the request. 

### Evidence:
[1] Britain (place name)
Britain, after which "Britain" became the more commonplace name for the island called Great Britain. After the Anglo-Saxon period, "Britain" was used as a historical term only. Geoffrey of Monmouth in his pseudohistorical "Historia Regum Britanniae" ...

[2] Great Britain
The peoples of these islands of "Prettanike" were called the "Priteni" or "Pretani". "Priteni" is the source of the Welsh language term Prydain, "Britain", which has the same source as the Goidelic term Cruithne used to refer to the early Brythonic-speaking inhabitants of Ireland. The latter were later called Picts or Caledonians ...

...

[10] Albion
Albion is an alternative name for Great Britain. The oldest attestation of the toponym comes from the Greek language. It is sometimes used poetically and generally to refer to the island, but is less common than 'Britain' today. The name for Scotland in most of the Celtic languages is related to Albion: "Alba" in Scottish Gaelic, "Albain" ...

### Instruction: What was Britain called - before it was Britain?

### Response:
\end{lstlisting}
\end{tcolorbox}
\caption{Prompt of Standard RAG for Non-instruction-tuned LM}
\label{fig:baseline-prompt}
\end{figure*}

\begin{figure*}[ht]
\begin{tcolorbox}[left=1mm,right=1mm,top=0.mm, bottom=0mm,colback=white]
\begin{lstlisting}[style=demo]
========================================= Prompt ========================================= 

[INST] Below is an instruction that describes a task. Write a response for it and state your explanation supporting your response.

### Instruction: What was Britain called - before it was Britain?
### Evidence:
[1] Britain (place name)
Britain, after which "Britain" became the more commonplace name for the island called Great Britain. After the Anglo-Saxon period, "Britain" was used as a historical term only. Geoffrey of Monmouth in his pseudohistorical "Historia Regum Britanniae" ...

[2] Great Britain
The peoples of these islands of "Prettanike" were called the "Priteni" or "Pretani". "Priteni" is the source of the Welsh language term Prydain, "Britain", which has the same source as the Goidelic term Cruithne used to refer to the early Brythonic-speaking inhabitants of Ireland. The latter were later called Picts or Caledonians ...

...

[10] Albion
Albion is an alternative name for Great Britain. The oldest attestation of the toponym comes from the Greek language. It is sometimes used poetically and generally to refer to the island, but is less common than 'Britain' today. The name for Scotland in most of the Celtic languages is related to Albion: "Alba" in Scottish Gaelic, "Albain" ...
[/INST] The response is:

\end{lstlisting}
\end{tcolorbox}
\caption{Prompt of Standard RAG for Instruction-tuned LM}
\label{fig:baseline-prompt-instruct}
\end{figure*}

\newpage

\section{Case Study}\label{appendix:case-study}
Figure~\ref{fig:case-study} and \ref{fig:case-study-2} are two representative cases from TriviaQA and PubHealth. They show the two drafts generated for the same question. We observe that our RAG drafter can well understand the multiple perspectives in the retrieval results and generate high-quality drafts. Our RAG verifier can also help filter out the unreliable drafts as we observe a relatively low scores in the first draft in Figure~\ref{fig:case-study} and \ref{fig:case-study-2}.

\begin{figure*}[ht]
\begin{tcolorbox}[left=1mm,right=1mm,top=0.mm, bottom=0mm,colback=white]
\begin{lstlisting}[style=demo]
=================================== Prompt of Drafter ==================================== 

Response to the instruction. Also provide rationale for your response.
## Instruction: Which actress/singer starred as Doralee Rhodes in the 1980 film, "Nine to Five"?
## Evidence: 
[1] Diana DeGarmo
Sheila until the show's closure on June 27, 2010. DeGarmo played the role of Doralee Rhodes in the national tour of "9 to 5", which was launched in Nashville on September 21, 2010. She ended her run as Doralee after the July 2011 Minneapolis tour stop. DeGarmo appeared on musical theatre composer Scott Alan's third album, ...
[2] 9 to 5 (musical)
Pippa Winslow as Violet Newstead, Louise Olley as Doralee Rhodes and Leo Sene as Franklin Hart Jr, with Samantha Giffard as Roz, Matthew Chase as Joe and Mark Houston, Rachel Ivy, and Blair Anderson. "9 to 5" will play in the West End at the Savoy Theatre from January 29 to August 31, 2019. The production stars Amber Davies (Judy), ...

====================================== Completion ========================================
## Rationale: Diana DeGarmo played the role of Doralee Rhodes in the national tour of "9 to 5", which began in September 2010.
## Response: Diana DeGarmo

====================================== Verification ====================================== 
draft score: 0.6594
self-containment score: 0.3417
self-reflection score: 0.5238

------------------------------------------------------------------------------------------

=================================== Prompt of Drafter ==================================== 

Response to the instruction. Also provide rationale for your response.
## Instruction: Which actress/singer starred as Doralee Rhodes in the 1980 film, "Nine to Five"?
## Evidence: 
[1] Cynthia Rhodes
Rhodes garnered her first non-dance related role in Michael Crichton's 1984 science fiction thriller "Runaway" with Tom Selleck, Kirstie Alley and Gene Simmons. Her most notable role was as dance instructor Penny Johnson in the hit 1987 motion picture "Dirty Dancing" with Jennifer Grey and Patrick Swayze. Rhodes' final motion picture role was the character of Vickie Phillips opposite actor Jameson Parker in the sleeper action-adventure movie ...
[2] Dolly Parton
Parton's feature film "9 to 5" (1980). The musical ran at the Ahmanson Theatre, Los Angeles in late 2008. It opened on Broadway at the Marquis Theatre in New York City, on April 30, 2009, to mixed reviews. The title track of her 2008 "Backwoods Barbie" album was written for the musical's character Doralee. Although her score (as well as the musical debut of actress Allison Janney) were praised, ...

====================================== Completion ========================================
## Rationale: Dolly Parton starred in the 1980 film "9 to 5" as Doralee Rhodes. Additionally, Parton's 2008 musical "9 to 5" was based on the film, with Parton reprising her role as Doralee.
## Response: Dolly Parton starred as Doralee Rhodes in the 1980 film, "Nine to Five".

====================================== Verification ====================================== 
draft score: 0.71
self-containment score: 0.4346
self-reflection score: 0.7449
\end{lstlisting}
\end{tcolorbox}
\caption{Case study of \method from TriviaQA where \textit{Dolly Parton} is the correct answer.}
\label{fig:case-study}
\end{figure*}

\newpage

\begin{figure*}[ht]
\begin{tcolorbox}[left=1mm,right=1mm,top=0.mm, bottom=0mm,colback=white]
\begin{lstlisting}[style=demo]
=================================== Prompt of Drafter ==================================== 

Response to the instruction. Also provide rationale for your response.
## Instruction: A lot of the problems with forest fires is because of bad policy (not to clear out the forests) that was put in place by environmentalists.
## Evidence: 
[1] Forest restoration
That can be because of provoked deforestation in order to substitute forests by crop areas, or in dry areas, because of wild fires occurring naturally or intentionally. A whole section of forest landscape restoration in linked to this particular problem, as in many cases, the net loss of ecosystem value is very high and can open the drop to an accelerated further degradation of the soil conditions through erosion and desertification. This indeed has dire consequences on both the quality of the habitats and their related fauna. ...
[2] Wildfire
Attempts to thin out the fuel load may be met with opposition due to the desirability of forested areas. Wildland goals may be further resisted because of endangered species protections and habitat preservation. The ecological benefit of fire is often overridden by the economic benefits of protecting structures and lives. Additionally, federal policies that cover wildland areas usually differ from local and state policies that govern urban lands. ...

====================================== Completion ========================================
## Rationale: Inadequate forest management can contribute to the severity of wildfires.
## Response: True

====================================== Verification ====================================== 
draft score: 0.8161
self-containment score:0.1360
self-reflection score: 0.8877

------------------------------------------------------------------------------------------

=================================== Prompt of Drafter ==================================== 

Response to the instruction. Also provide rationale for your response.
## Instruction: A lot of the problems with forest fires is because of bad policy (not to clear out the forests) that was put in place by environmentalists.
## Evidence: 
[1] 1997 Indonesian forest fires
The 1997 Indonesian forest fires were caused by changing land use which made the tropical forest vulnerable to fire during a drought associated with that year's El Nino. Indonesian forests have historically been resistant to burning even during long dry seasons and despite the use of fire to clear land for swidden agriculture. The land use changes that led to the fires were a combination of industrial-scale logging, draining peatlands for conversion to oil palm and fast-growing tree plantations, and a massive government program to drain swamps and convert them to rice paddies. ...
[2] Wildfire
Attempts to thin out the fuel load may be met with opposition due to the desirability of forested areas. Wildland goals may be further resisted because of endangered species protections and habitat preservation. The ecological benefit of fire is often overridden by the economic benefits of protecting structures and lives. Additionally, federal policies that cover wildland areas usually differ from local and state policies that govern urban lands. ...

====================================== Completion ========================================
## Rationale: The 1997 Indonesian forest fires were caused by **land use changes**, including industrial-scale logging, draining peatlands.
## Response: False

====================================== Verification ====================================== 
draft score: 0.8288
self-containment score: 0.1325
self-reflection score: 0.9533
\end{lstlisting}
\end{tcolorbox}
\caption{Case study of \method from PubHealth where \textit{False} is the correct answer.}
\label{fig:case-study-2}
\end{figure*}

\end{document}